\documentclass[10pt,twocolumn,letterpaper]{article}

\usepackage[pagenumbers]{cvpr} 

\usepackage{multirow}
\usepackage{arydshln}
\usepackage{placeins}
\usepackage{cuted}

\usepackage{algorithm}
\usepackage{algpseudocode}

\newcommand{\method}{GenMTL\xspace}

\definecolor{cvprblue}{rgb}{0.21,0.49,0.74}
\usepackage[pagebackref,breaklinks,colorlinks,allcolors=cvprblue]{hyperref}

\def\confName{CVPR}
\def\confYear{2027}

\title{Generative Models are Generalist Multi-Task Dense Predictors}

\author{
Yangyang Xu$^{1}$ \quad
Yingjie Wang$^{2}$ \quad
Xi Ye$^{1}$ \quad
Jun Zhu$^{1}$\thanks{Corresponding author}\\[-2pt]
{\normalsize$^{1}$Tsinghua University,  \quad $^{2}$Beijing University of Posts and Telecommunications}
}

\makeatletter
\patchcmd{\@maketitle}
    {\vspace*{24pt}}
    {\vspace*{8pt}}{}{}
\makeatother

\begin{document}
\maketitle

\begin{abstract}
Transferring pretrained video generators to dense perception involves a central design choice: how to expose their learned representations to multi-task dense predictions.
A common approach is to share not only the generative backbone but also its synthesis output space, casting semantic segmentation, metric depth, surface normals, and boundary detection as a unified reconstruction problem. 
We investigate how task-specific prediction heads can jointly use the internal transformer features and output latents of a shared pretrained video generator.
Motivated by this observation, we introduce \method{}, a framework that adapts a pretrained video generator into a generalist dense predictor through supervised task-specific adaptation.
Each task is equipped with a lightweight low-rank adapter and a task-specific prediction head, enabling diverse dense outputs without forcing them into a common generative space. 
Each prediction head fuses pre-projection transformer features with the predicted latent and produces task-specific dense outputs.
Experiments on NYUD-v2, PASCAL-Context, and Cityscapes demonstrate the effectiveness of our framework across diverse dense prediction tasks. These results highlight the potential of pretrained video generators for multi-task scene understanding.
We believe our approach advances the use of pretrained video generators for multi-task dense predictions.

\end{abstract}

\section{Introduction}

Rapid advances in image and video generation have made generative pretraining a promising foundation for visual perception.
Vision Banana represents task outputs as RGB images, enabling a pretrained image generator to support semantic and geometric tasks~\citep{visionbanana2026}.
GenCeption adapts a pretrained video generator into an instruction-conditioned, feed-forward perception model for a broad range of vision tasks~\citep{genception2026}.
These advances motivate a closer examination of how representations learned for video generation should be used for multi-task dense prediction.

For multi-task dense prediction, a key design choice is how to connect a shared generator to task-specific outputs. GenCeption represents dense outputs in a common RGB format, while StableMTL uses a unified target-latent objective under partial supervision~\citep{genception2026,stablemtl2026}. 
Sharing pretrained weights does not require encoding all task targets as RGB images or latents.
Task-specific decoders and supervision, as explored by GenPercept~\citep{xu2024genpercept}, provide an alternative. With such decoders, a further design choice concerns their inputs: the internal transformer features before the final synthesis projection, the predicted latent produced by that projection, or both. We investigate how combining these representations can support multiple dense tasks while preserving task-specific prediction.

We introduce \method, which adapts a shared video generator for multi-task dense prediction through feature--latent fusion and task-specific prediction.
Each fusion module combines normalized transformer features immediately before the synthesis projection with the predicted latent produced by the same diffusion transformer (DiT) forward pass.
In this design, the synthesis projection serves as a pretrained feature transformation within the perception pipeline.
Each prediction head maps the fused features directly to the required task output and is trained with the corresponding task loss.

To support multiple tasks within this design, we use a shared, frozen Wan2.1-T2V-14B backbone and equip each task with a lightweight LoRA adapter and prediction head~\citep{wan2025,hu2022lora}.
A fixed task instruction conditions the backbone, while the known task identity selects the corresponding adapter and head.
Task-balanced training updates the adapters and heads while retaining one shared pretrained weights.
Following GenCeption's feed-forward formulation~\citep{genception2026}, inference requires one deterministic transformer evaluation per requested task, followed by direct dense prediction without iterative denoising or VAE decoding.

Experiments on NYUD-v2~\citep{silberman2012nyud}, PASCAL-Context~\citep{pascal2014}, and Cityscapes~\citep{cityscapes} show that \method{} achieves the best performance among the compared methods on several dense prediction tasks.
These results are obtained by training only task-specific adapters, fusion modules, and prediction heads while keeping the shared Wan2.1-T2V-14B backbone frozen.
With one checkpoint per benchmark, \method{} supports four tasks on NYUD-v2, five on PASCAL-Context, and two on Cityscapes, spanning semantic, geometric, and boundary prediction.
Our contributions are as follows:
\begin{itemize}
    \item We introduce \method, a framework for multi-task dense prediction that shares a frozen pretrained video generator while learning task-specific adapters, fusion modules, and prediction heads.

    \item  We design feature--latent fusion within task-routed DiT adaptation to combine pre-projection transformer features with predicted latents, retaining the pretrained synthesis transformation as part of direct task prediction.

    \item Experiments on NYUD-v2, PASCAL-Context, and Cityscapes demonstrate the effectiveness of \method, with the best performance among the compared methods on several tasks. Ablations further examine adapter sharing, LoRA rank, and task-specific text conditioning. 
\end{itemize}

\section{Related Work}
\label{sec:related_work}

\paragraph{Multi-task learning.}
Dense multi-task learning jointly learns related pixel-level tasks to improve parameter reuse and information sharing.
Early approaches learn where and how to share representations through cross-stitch units, layer-wise feature fusion, task-specific attention, and task-relational context~\citep{misra2016crossstitch,gao2019nddrcnn,liu2019mtan,bruggemann2021atrc}.
PAD-Net distills information across task predictions, while MTI-Net models task interactions at multiple scales~\citep{xu2018padnet,vandenhende2020mtinet}.
Transformer-based methods extend this approach: InvPT performs spatial and cross-task reasoning with an inverted pyramid, MQTransformer uses task queries to retrieve relevant features, and TaskPrompter injects learned prompts into shared representations~\citep{ye2022invpt,xu2022mqtransformer,ye2023taskprompter}.
TSP-Transformer uses task-specific prompts to balance shared representations and task-specific capacity~\citep{wang2024tsptransformer}, while PGT conditions dense prediction on task prompts~\citep{lu2023pgt}.
Recent methods emphasize dynamic task-specific representations: DeMT combines deformable feature mixing with task-aware decoding, TaskExpert uses a memorial mixture-of-experts, and MLoRE combines low-rank experts with a shared generic path~\citep{xu2023demt,ye2023taskexpert,yang2024mlore}.
Other directions explore cross-task distillation, parameter-efficient task-agnostic and task-specific adaptation, and joint-task regularization under partial labels~\citep{xu2023mtkd,agiza2024mtlora,nishi2024jointtask}.
DiffusionMTL uses joint diffusion and denoising with multi-task conditioning for partially annotated tasks~\citep{ye2024diffusionmtl}.
TaskDiffusion models task relations through joint denoising and a cross-task diffusion decoder~\citep{yang2025taskdiffusion}.
These methods explore feature sharing, task-conditioned features, and task interactions. \method{} studies how a pretrained video generator can provide shared representations while retaining task-specific adaptation and prediction.

\paragraph{Generalist visual prediction.}
Generalist vision models pursue broad task coverage through a shared model and a reusable prediction interface. Sequence formulations such as Pix2Seq express visual outputs as discrete tokens, while Unified-IO and Unified-IO 2 progressively extend token-based unified modeling across vision, language, audio, action, and other multimodal tasks~\citep{chen2022pix2seq,lu2022unifiedio,lu2024unifiedio2}. Related sequence-to-sequence systems unify tasks through prompted pixel-to-sequence interfaces or common formulations spanning vision and language under a single parameterization~\citep{chen2022pix2seqv2,wang2022ofa,wang2023painter}. 
Other generalist designs combine learned guiding codes, task-agnostic decoding, or joint pixel-and-language decoding within shared architectures for diverse perception tasks~\citep{kolesnikov2022uvim,li2023uniperceiverv2,zou2023xdecoder}. 
Vision Banana adapts a pretrained image generator to diverse vision tasks by representing their outputs as RGB images~\citep{visionbanana2026}.
These methods show that a shared model can support multiple vision tasks. However, sharing pretrained weights does not require encoding all task targets as RGB images or latents. \method{} uses a shared pretrained backbone conditioned on task instructions, with a separate prediction head for each task to directly produce the required output.

\paragraph{Repurposing generative models for dense predictions.}
Generative pretraining provides representations that can be reused for perception. Features from diffusion models have supported semantic segmentation, correspondence, and text-conditioned dense prediction~\citep{baranchuk2022labels,wang2023dift,zhang2023vpd}. These approaches extract or adapt internal representations rather than treating the generator itself as the prediction interface. Conditional diffusion can instead operate directly as a noise-to-map predictor for dense outputs such as segmentation and depth~\citep{ji2023ddp}. Task-specific adaptations further demonstrate the utility of generative priors for geometry: Marigold repurposes a pretrained diffusion image generator for monocular depth, GeoWizard jointly estimates depth and normals, while Lotus adopts direct annotation prediction and a one-step formulation for both tasks~\citep{ke2024marigold,fu2024geowizard,he2025lotus}. More direct repurposing is explored by GenPercept, which converts an image diffusion model into a deterministic one-step predictor for several dense tasks, including depth, normals, segmentation, and matting~\citep{xu2024genpercept}. Recent work further studies how diffusion-based perception scales across multiple tasks and training and inference compute regimes~\citep{ravishankar2025scaling}. StableMTL formulates multi-task dense prediction as task-conditioned latent regression with a pretrained latent diffusion model, and studies learning heterogeneous tasks from synthetic datasets that provide only partial task annotations~\citep{stablemtl2026}. Video diffusion models have also been extended toward joint generation, entity segmentation, and depth prediction~\citep{yang2025unifieddensevideo}. GenCeption further uses a pretrained text-to-video diffusion backbone as an instruction-steered feed-forward perception model~\citep{genception2026}. Building on this generative setting, \method{} examines whether heterogeneous dense tasks should also share a synthesis-oriented output representation. It retains task-conditioned generative representations while using task-specific native-space readouts matched to distinct output structures.

\begin{figure*}[!t!]
\centering
  \includegraphics[width=0.9999\textwidth]{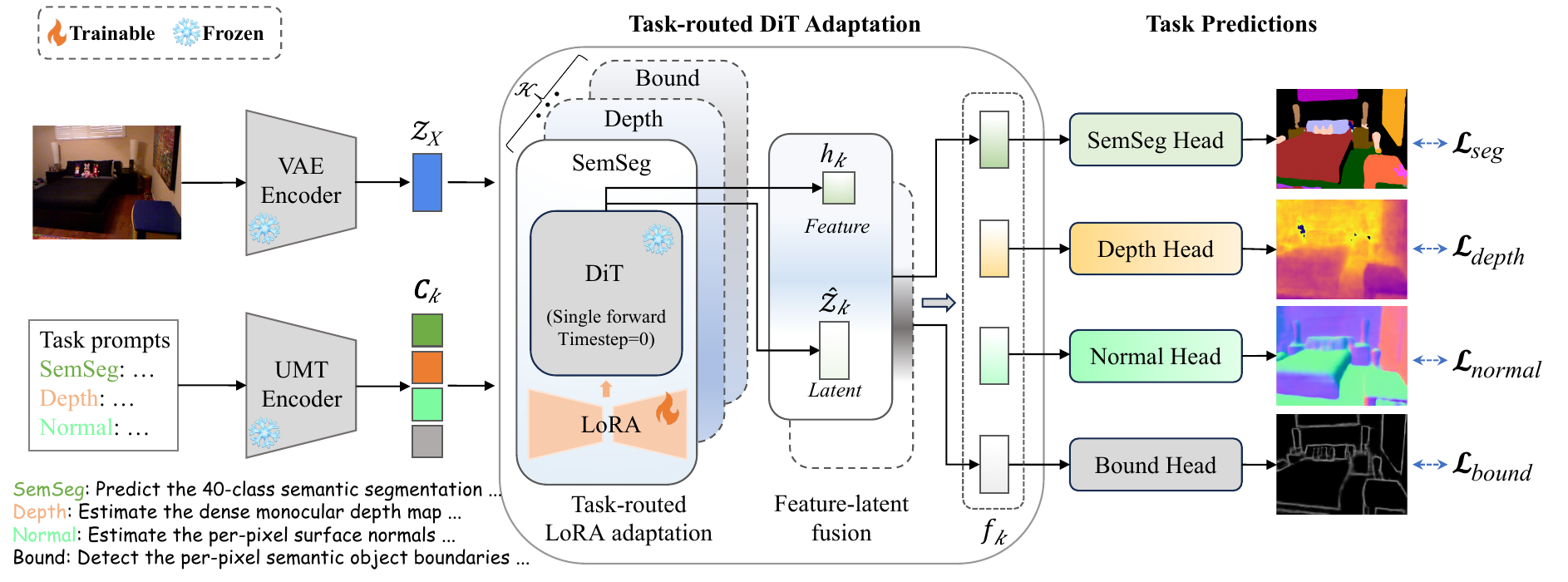}
 \caption{Architecture overview of GenMTL, a task-routed framework for dense multi-task prediction adapted from a pretrained video diffusion model. Given an input image and a text prompt specifying the target task, GenMTL activates the corresponding LoRA adapter within a shared, frozen diffusion transformer (DiT) backbone. Task-routed DiT adaptation consists of task-routed LoRA adaptation and feature--latent fusion. 
 A feature-latent fusion module combines pre-projection transformer features with the predicted latent, and a prediction head maps the fused features to dense outputs.
 }
\label{fig:over_view}
\end{figure*}

\section{Method}
\label{sec:method}

In this section, we present \method, a framework that adapts a shared video generator to mu lti-task dense predictions. We detail the two key designs in \method: task-routed DiT adaptation and task predictions in Sec.~\ref{sec:GenMTL}, followed by the multi-task training objectives and inference in Sec.~\ref{sec:losses}. An overview of \method is presented in Figure~\ref{fig:over_view}.

\subsection{Formulation}
\label{sec:formulation}
We denote the RGB input to the VAE encoder as $x\in\mathbb{R}^{B\times3\times T\times H\times W}$, where $B$ is the batch size, $H$ and $W$ are the spatial dimensions, and $T=1$ for the images considered in our experiments.
Let $\mathcal K$ ( $\mathcal K=\{\mathrm{SemSeg}, \mathrm{PartSeg}, \mathrm{Depth},\mathrm{Normal},\mathrm{Sal},\mathrm{Edge}\}$) denote the set of dense prediction tasks.
For each task $k\in\mathcal K$, the training data provide a spatially aligned target $y_k$ and a validity mask $\Omega_k$.
The targets occupy different native domains: categorical labels for semantic and human-part segmentation, binary labels for saliency and boundaries, positive metric values for depth, and three-dimensional unit vectors for surface normals.
A fixed natural-language instruction $p_k$ specifies the requested task.

Our predictor shares the pretrained video generator while using a task-specific LoRA adapter and prediction head for each $k$.
Denoting the frozen video VAE encoder by $E$ and the frozen text encoder by $T_{\mathrm{text}}$, we formulate the prediction pipeline as follows:
\begin{equation}
\begin{aligned}
    z_x &= E(x), \quad c_k=T_{\mathrm{text}}(p_k),\\
    (\hat z_k,h_k) &= F_{\theta_0,\Delta\theta_k}(z_x,c_k;t=0),\\
    f_k &= M_{\eta_k}(\hat z_k,h_k),\\
s_k &= D_{\psi_k}(f_k),  \\
    \hat y_k &= \rho_k\!\left(\operatorname{Resize}_k(s_k)\right).
\end{aligned}
\label{eq:formulation}
\end{equation}
Here $F_{\theta_0,\Delta\theta_k}$ denotes the DiT adapted for task $k$.
$\theta_0$ denotes the frozen pretrained DiT parameters, and $\Delta\theta_k$ denotes the task-specific LoRA parameters selected by the known task identity.
The DiT provides the pre-projection transformer features $h_k$ and the predicted latent $\hat z_k$, defined as its negated velocity output.
$D_{\psi_k}$ is the task-specific prediction head, and $s_k$ is its raw output.
$\operatorname{Resize}_k$ resizes this output to the annotation grid and is the identity when the resolutions match.
The mapping $\rho_k$ then retains categorical and boundary logits, makes depth positive, or normalizes surface normals to unit length.
The trainable parameters are $\Phi=\{(\Delta\theta_k,\eta_k,\psi_k)\}_{k\in\mathcal K}$, optimized with the objective in Eq.~\ref{eq:mtl_objective}.
Inference uses one deterministic DiT forward pass per requested task; requesting all tasks uses $|\mathcal K|$ passes with different active adapters.

\subsection{GenMTL architecture}
\label{sec:GenMTL}
We use the VAE encoder, text encoder, and DiT of Wan2.1-T2V-14B~\citep{wan2025}, keeping all pretrained parameters frozen.
Generative models have two distinct interfaces for perception: the representation formed by the transformer and the low-dimensional latent produced for synthesis.
GenMTL treats the former as shared pretrained knowledge and the latter as only one of several signals available to a dense predictor.
For a requested task $k$, the known task identity selects a text instruction, a lightweight adaptation of the common transformer, and a head that predicts in the task's native output space.
This separates \emph{what is shared} across tasks from \emph{how each task reads it}, without forcing categorical and geometric targets into a common RGB reconstruction objective. Task-routed DiT adaptation consists of two key components: task-routed LoRA adaptation and feature--latent fusion, as illustrated in Figure~\ref{fig:over_view}.
LoRA adapts the attention layers within the DiT, while feature--latent fusion combines its pre-projection transformer features and predicted latent for the task-specific prediction head.

\subsection{Task-routed DiT adaptation}
\label{sec:task_routed_adaptation}

\paragraph{Task-routed LoRA adaptation.}
We adapt the shared DiT to each task through low-rank updates to its attention layers.
For a request $k$, the known task identity activates the adapter $\Delta\theta_k$ and its associated prediction head $D_{\psi_k}$, while the instruction embedding $c_k$ conditions the transformer.
The same task adapter is active throughout the DiT; routing requires neither an input-dependent gating network nor a mixture of task adapters.
Thus, tasks share pretrained weights while retaining task-dependent transformations of the input.

We implement $\Delta\theta_k$ using LoRA~\citep{hu2022lora} on the query, key, value, and output projections of both self-attention and cross-attention.
Let $\mathcal J$ index these projections across all DiT blocks.
For each frozen weight $W_0^{(j)}\in\mathbb R^{d_{\mathrm{out}}^{(j)}\times d_{\mathrm{in}}^{(j)}}$, task $k$ learns factors $A_k^{(j)}\in\mathbb R^{r\times d_{\mathrm{in}}^{(j)}}$ and $B_k^{(j)}\in\mathbb R^{d_{\mathrm{out}}^{(j)}\times r}$, giving
\begin{equation}
\begin{aligned}
    W_k^{(j)} &= W_0^{(j)}+\frac{\alpha}{r}B_k^{(j)}A_k^{(j)},\\
    W_k^{(j)}u &= W_0^{(j)}u+\frac{\alpha}{r}B_k^{(j)}\!\left(A_k^{(j)}u\right),
    \qquad j\in\mathcal J,
\end{aligned}
\label{eq:task_lora}
\end{equation}
where $u$ is an input activation, $r$ is the adapter rank, and $\alpha$ controls the update scale; pretrained biases are unchanged and omitted for clarity.
$j$ indexes the attention projection layers equipped with LoRA.
We write $\Delta\theta_k=\{A_k^{(j)},B_k^{(j)}\}_{j\in\mathcal J}$ for the collection of task-specific factors.
All other DiT parameters are frozen, including the feed-forward layers and conditioning modules.

Given the encoded image $z_x$ and instruction $c_k$, a frozen patch embedding converts $z_x$ into $N$ visual tokens of width $d$, and a frozen text projection maps $c_k$ to the cross-attention context.
Following GenCeption~\citep{genception2026}, we use the clean input latent without adding noise and fix the timestep to $t=0$.
The forward computation through $L$ DiT blocks is
\begin{equation}
\begin{aligned}
    H_k^{(0)} &= \operatorname{Patch}_0(z_x),
    \qquad \bar c_k=P_0^{\mathrm{text}}(c_k),\\
    H_k^{(\ell)} &= \mathcal B_{\theta_0,\Delta\theta_k}^{(\ell)}
    \!\left(H_k^{(\ell-1)},\bar c_k;t=0\right),
    \quad \ell=1,\ldots,L,
\end{aligned}
\label{eq:routed_dit_blocks}
\end{equation}
where $\operatorname{Patch}_0$ includes patch embedding and token flattening, $H_k^{(\ell)}\in\mathbb R^{B\times N\times d}$, and $\mathcal B^{(\ell)}$ denotes the $\ell$-th pretrained block with the selected low-rank updates.
Each block applies visual self-attention, text-conditioned cross-attention, and a feed-forward transformation in sequence, retaining the pretrained normalization, residual connections, and timestep-dependent modulation.
Self-attention derives queries, keys, and values from the visual tokens; cross-attention derives queries from the visual tokens and keys and values from $\bar c_k$.
Both attention operations use the adapted projections in Eq.~\ref{eq:task_lora}, while preserving the pretrained query--key normalization.

After the final block, we retain the normalized and timestep-modulated tokens immediately before the generative output projection:
\begin{equation}
\begin{aligned}
    \tilde H_k &= \operatorname{LN}_0\!\left(H_k^{(L)}\right)
        \odot\bigl(1+a_0(0)\bigr)+b_0(0),\\
    h_k &= \mathcal G(\tilde H_k),\\
    \hat z_k &= -\mathcal U\!\left(P_0(\tilde H_k)\right)
              =-v_{\theta_0,\Delta\theta_k}(z_x,0,c_k).
\end{aligned}
\label{eq:one_step}
\end{equation}
Here $a_0(0)$ and $b_0(0)$ are the frozen output-modulation scale and shift at $t=0$, and $\odot$ denotes element-wise multiplication.
$\mathcal G$ reshapes tokens into the transformer patch grid, whereas $\mathcal U$ unpatchifies the output of the frozen projection $P_0$ into the VAE latent grid.
For the 14B backbone, $h_k$ contains 5120-channel features on the transformer patch grid, while $\hat z_k$ has 16 channels on the VAE latent grid.
These operations define the pair $(\hat z_k,h_k)=F_{\theta_0,\Delta\theta_k}(z_x,c_k;0)$ in Eq.~\ref{eq:formulation}.
Both representations are passed to $D_{\psi_k}$, without iterative denoising or VAE decoding of predictions.
The task-specific loss propagates through both branches to the selected adapter and head, with pretrained parameters frozen.

\paragraph{Feature--latent fusion.}
\label{sec:feature_latent_fusion}
Feature--latent fusion gives the prediction head access to both the pre-projection transformer features and their transformation through the frozen synthesis projection.
The two representations differ in channel width and spatial resolution, so we first project them to 256 channels and align them on the latent grid at $H/8\times W/8$.
Each task has its own fusion parameters $\eta_k$.
For $T=1$, we fold the singleton frame dimension into the batch dimension for 2D processing and compute
\begin{equation}
\begin{aligned}
    \tilde h_k &= \mathcal A_k\!\left(\Pi^h_k(h_k)\right),
    &\tilde z_k &= \Pi^z_k(\hat z_k),\\
    f_k^0 &= \operatorname{GELU}(\tilde h_k+\tilde z_k),
    &f_k &= f_k^0+\mathcal R_k(f_k^0).
\end{aligned}
\label{eq:additive_fusion}
\end{equation}
Here $\Pi^h_k$ and $\Pi^z_k$ project the channels with normalization, $\mathcal A_k$ upsamples and refines the feature branch, and $\mathcal R_k$ is a depthwise--pointwise residual refinement block.
The resulting $f_k=M_{\eta_k}(\hat z_k,h_k)$ is the output of task-routed DiT adaptation and the input to the prediction head.
The task loss propagates through both fusion branches to the selected LoRA adapter.
Appendix~\ref{app:framework} gives the layer configuration.

\paragraph{Task predictions.}
The prediction heads convert the fused features into the outputs required by each task.
Each head $D_{\psi_k}$ applies three $2\times$ upsampling stages to recover the encoder input resolution, followed by a task-specific projection that produces the raw output $s_k$.
Appendix~\ref{app:framework} gives the layer configuration and tensor dimensions.

\noindent
\textbf{Task-specific outputs.}
We resize the head output $s_k$ to the resolution of the annotation.
Categorical and boundary predictions remain as logits, while depth is constrained to be positive and surface normals are normalized:
\begin{equation}
    \rho_k(s)=
    \begin{cases}
        s,&\text{categorical or boundary logits},\\
        \operatorname{softplus}(s)+\epsilon,&\text{depth},\\
        s/\max(\lVert s\rVert_2,\epsilon),&\text{surface normals}.
    \end{cases}
    \label{eq:native_prediction}
\end{equation}
The norm is computed over the three normal components, with a small constant $\epsilon>0$ to avoid division by zero.
Saliency uses two categorical logits, and boundary detection uses one logit per pixel.
These outputs are supervised directly with the task annotations; categorical and boundary logits are converted to labels or probabilities for evaluation.

\subsection{Training and inference}
\label{sec:losses}
\textbf{Training.} We supervise the final task predictions $\hat y_k$ in Eq.~\ref{eq:formulation} using the corresponding annotations.
The predicted latent is used in feature--latent fusion, without a target-latent reconstruction loss.
For each optimization step, we minimize the mean task loss:
\begin{equation}
    \mathcal L_{\mathrm{MTL}}(\Phi)
    = \frac{1}{|\mathcal K|}
      \sum_{k\in\mathcal K}
      \ell_k(\hat y_k,y_k;\Omega_k),
    \label{eq:mtl_objective}
\end{equation}
where $\ell_k$ is the weighted loss for task $k$ over its training batch, and $\Omega_k$ restricts supervision to valid pixels.
Categorical tasks use cross-entropy-based losses.
Depth and surface normals use dataset-specific regression losses, while boundaries use class-balanced binary cross entropy. 
The loss definitions and weights are provided in Appendix~\ref{app:task_losses}.
Each task loss updates the corresponding LoRA adapter, fusion module, and prediction head.

\noindent
\textbf{Inference.}
For a requested task, we select its instruction, LoRA adapter, fusion module, and prediction head.
The input latent passes through the DiT once at $t=0$, followed by feature--latent fusion and direct task prediction.
Predicting all tasks requires $|\mathcal K|$ DiT forward passes, without iterative denoising or VAE decoding of the predictions.

\section{Experiments}
\label{sec:experiments}

We evaluate whether a pretrained video generator can serve as a generalist backbone for dense prediction after supervised task-specific adaptation.
For each benchmark, we learn a separate model whose tasks share the frozen generator and use their own LoRA adapters, feature--latent fusion modules, and prediction heads.
The main experiments assess this formulation across semantic, geometric, and boundary tasks, while ablations examine adapter sharing, adaptation capacity, and text conditioning.

\begin{table*}[!t]
\centering
\small
\caption{Results on PASCAL-Context. ST baseline denotes the single-task baseline. $\Delta_m$ measures the average relative multi-task performance change.
$^\dagger$\method{} trains 270M parameters while keeping the pretrained Wan2.1-T2V-14B backbone and VAE frozen.
\textbf{Bold}/\underline{underlined}: best/second best; $\uparrow$/$\downarrow$: higher/lower is better.}
\label{tab:sota_comparison_pascal}
\begin{tabular}{lcccccccc}
\toprule
\multirow{2}*{Model} &\multirow{2}*{Backbone}
& {SemSeg}
& {PartSeg}
& {Sal}
& {Normal}
& {Bound}
& {$\Delta_m$}
& \shortstack{Trainable Params}\\
&
& (mIoU)$\uparrow$
& (mIoU)$\uparrow$
& (maxF)$\uparrow$
& (mErr)$\downarrow$
& (odsF)$\uparrow$
& $ [\%]$$\uparrow$
& (M)$\downarrow$\\
\midrule
ST baseline                                   &ViT-L &81.62 &72.21 &84.34 &13.59 &\underline{76.79} &0.00 &317\\
PAD-Net~\citep{xu2018padnet}                  &ViT-L &78.01 &67.12 &79.21 &14.37 &72.60 &-5.72 &330\\
MTI-Net~\citep{vandenhende2020mtinet}         &ViT-L &78.31 &67.40 &84.75 &14.67 &73.00 &-4.62 &851\\
ATRC~\citep{bruggemann2021atrc}               &ViT-L &77.11 &66.84 &81.20 &14.23 &72.10 &-5.50 &340\\
MQTransformer~\citep{xu2022mqtransformer}     &ViT-L &77.72 &65.14 &84.43 &14.63 &54.77 &-10.16 &314\\
DeMT~\citep{xu2023demt}                       &ViT-L &78.96 &67.39 &84.26 &14.53 &55.29 &-8.99 &308\\
InvPT~\citep{ye2022invpt}                     &ViT-L &79.03 &67.61 &84.81 &14.15 &73.00 &-3.61 &423\\
TaskPrompter~\citep{ye2023taskprompter}       &ViT-L &80.89 &68.89 &84.83 &13.72 &73.50 &-2.03 &401\\
TaskExpert~\citep{ye2023taskexpert}           &ViT-L &80.64 &69.42 &84.87 &13.56 &73.30 &-1.74 &420\\
DiffusionMTL~\citep{ye2024diffusionmtl}       &ViT-L &80.46 &69.13 &84.85 &14.02 &70.96 &-3.16 &381\\
TSP-Transformer~\citep{wang2024tsptransformer}&ViT-L &81.48 &70.64 &84.86 &13.69 &74.80 &-1.01 &422\\
MLoRE~\citep{yang2024mlore}                   &ViT-L &81.41 &70.52 &84.90 &{13.51} &75.42 &-0.64 &407\\
TaskDiffusion~\citep{yang2025taskdiffusion}   &ViT-L &81.58 &71.30 &\underline{85.05} &\textbf{13.43} &76.07 &-0.04 &472\\
SAK~\citep{lu2025sak}                         &ViT-L &\underline{84.01} &\underline{76.99} &84.65 &13.82 &76.27 &\underline{2.30}  &407\\
\hdashline
ST baseline                                   &Wan2.1 &60.44 &68.41 &82.93 &15.25 &76.27 &0.00  &\textbf{7.96}\\
\method                                       &Wan2.1 &\textbf{84.56} &\textbf{80.59} &\textbf{85.21} &\underline{13.50} &\textbf{79.10} &\textbf{14.86} &\underline{270}$^\dagger$\\
\bottomrule
\end{tabular}
\end{table*}

\subsection{Experimental setup}
\label{sec:experimental_setup}
\textbf{Datasets.}
We evaluate GenMTL on PASCAL-Context~\citep{pascal2014}, NYUD-v2~\citep{silberman2012nyud}, and Cityscapes~\citep{cityscapes} under their respective multi-task protocols.
PASCAL-Context contains 4,998 training and 5,105 validation images for semantic segmentation, human-part segmentation, saliency detection, surface-normal estimation, and semantic-boundary detection.
Semantic and human-part segmentation use 21 and 7 classes, respectively, both including background.
NYUD-v2 comprises 795 training and 654 validation images, supporting 40-class semantic segmentation, monocular depth estimation, surface-normal estimation, and semantic-boundary detection.
Following the MTSAM task setup~\citep{wang2025mtsam}, we evaluate joint seven-class semantic segmentation and depth prediction in cityscapes.

\noindent
\textbf{Metrics.}
On PASCAL-Context~\citep{pascal2014}, we report mIoU for semantic and human-part segmentation, maximum F-score (maxF) for saliency, mean angular error (mErr) for surface normals, and optimal-dataset-scale F-score (odsF) for boundaries.
On NYUD-v2~\citep{silberman2012nyud}, we report mIoU for semantic segmentation, root mean squared error (RMSE) for depth, mean angular error (mErr) for surface normals, and optimal-dataset-scale F-score (odsF) for boundaries.
On Cityscapes~\citep{cityscapes}, we report mIoU and pixel accuracy (Pix Acc) for semantic segmentation, and absolute error (Abs Err) and relative error (Rel Err) for depth.
Where matched single-task references under a compatible evaluation setting are available, we additionally report the average relative multi-task performance
    $\Delta_m=\frac{1}{\mathcal{T}}\sum_{t=1}^{\mathcal{T}}
    (-1)^{l_t}\frac{F_{m,t}-F_{s,t}}{F_{s,t}}\times100\%,$
where $F_{m,t}$ and $F_{s,t}$ denote multi-task and single-task performance, respectively.

\noindent
\textbf{Implementation.}
We build \method{} on Wan2.1-T2V-14B~\citep{wan2025}, while keeping the pretrained VAE, UMT5 encoder, and original DiT backbone frozen. For NYUD-v2, we train a freshly initialized rank-32 LoRA adapter together with four native feature--latent prediction heads on eight GPUs, using a 500-step warm-up and polynomial learning-rate decay over at most 40k optimization steps. Table~\ref{tab:stoa_nyud_moe} reports the 20k-step checkpoint, evaluated at $448\times576$ resolution; its native depth head predicts in metric space without scale alignment. For PASCAL-Context, we use five routed rank-16 task-specific LoRA adapters and five native feature--latent prediction heads, trained on four GPUs with one example per rank and five gradient-accumulation steps, a 500-step warm-up, and polynomial decay over 40k optimizer steps; the adapters are frozen for the first 1,000 steps.

\subsection{Main results}

\noindent
\textbf{PASCAL-Context performance.}
Table~\ref{tab:sota_comparison_pascal} shows that \method{} achieves the best results on four of the five tasks.
Its strongest gains over prior multi-task methods are in human-part segmentation and boundary detection, surpassing SAK~\citep{lu2025sak} by 3.60 mIoU and 2.83 odsF points, respectively.
Semantic segmentation and saliency detection also achieve the highest scores of 84.56 mIoU and 85.21 maxF.
For surface normals, \method{} obtains 13.50 mErr, remaining competitive with TaskDiffusion's 13.43.
Compared with the Wan2.1 single-task baseline, \method{} improves all five task metrics, yielding an average relative gain of 14.86\%.
The model trains 270M parameters, fewer than the compared multi-task methods, while keeping the pretrained backbone frozen.
This combination of broad task coverage and leading accuracy supports the use of a shared video generator as a generalist dense prediction backbone.

\noindent
\textbf{NYUD-v2 performance.}
Table~\ref{tab:stoa_nyud_moe} shows that \method{} achieves the best performance among the compared methods on all four tasks. The gains over SAK~\citep{lu2025sak}, the strongest competing method in the table, are particularly pronounced in geometric prediction: depth RMSE decreases by 11.55\%, and surface-normal error decreases by 6.28\%. Boundary detection also improves by 1.92 odsF points, while semantic segmentation slightly increases from 63.18 to 63.25 mIoU. Relative to the Wan2.1 ST baseline, \method{} improves all four task metrics, including a + 4.73 mIoU gain in segmentation and a 12.52\% reduction in depth, yielding an average relative improvement of approximately 7.4\%. 
These results show that task-specific adaptation of a shared generator supports leading semantic accuracy together with improved geometric estimation and boundary localization.

\begin{table}[!t]
\centering
\small
\setlength{\tabcolsep}{1.5pt}
\setlength{\belowcaptionskip}{4pt}
\caption{Results on NYUD-v2.
$^\dagger$ViT-L single-task baseline.
$^\ddagger$Wan2.1-T2V-14B single-task baseline (backbone and VAE frozen), the $\Delta_m$ reference for \method{}.
\textbf{Bold}/\underline{underlined}: best/second best; $\uparrow$/$\downarrow$: higher/lower is better.}
\label{tab:stoa_nyud_moe}
\begin{tabular}{lcccccc}
\toprule
 \multirow{2}*{Model}   &SemSeg &Depth & Normal & Bound &{$\Delta_m$}\\
        &(mIoU)$\uparrow$   &(rmse)$\downarrow$   & (mErr)$\downarrow$  &(odsF)$\uparrow$ &[\%]$\uparrow$\\
\midrule
ST baseline$^\dagger$                           &56.77 &0.5141 &18.56 &78.93 &0.00\\
InvPT~\citep{ye2022invpt}                       &53.56 &0.5183 &19.04 &78.10 &-2.52\\
Task Prompter~\citep{ye2023taskprompter}        &55.30 &0.5152 &18.47 &78.20 &-0.81\\
TaskExpert~\citep{ye2023taskexpert}             &55.35 &0.5157 &18.54 &78.40 &-0.84\\
TSP-Transformer~\citep{wang2024tsptransformer}  &55.39 &0.4961 &18.44 &77.5 &-0.02\\
MLoRE~\citep{yang2024mlore}                     &55.96 &0.5076 &18.33 &78.43 &0.11\\
Task Diffusion~\citep{yang2025taskdiffusion}    &55.65 &0.5020 &18.43 &78.64 &0.18\\
 SAK~\citep{lu2025sak}                          &\underline{63.18} &\underline{0.4313} &16.25 &\underline{79.43} &\textbf{14.05}\\
\hdashline
ST baseline$^\ddagger$ (Wan2.1)                 &58.52 &0.4361 &\underline{16.21} &79.10 &0.00\\
{\method (ours)}                                &\textbf{63.25} &\textbf{0.3815} &\textbf{15.23} &\textbf{82.40} &\underline{7.39}\\
\bottomrule
\end{tabular}
\end{table}

\begin{table*}[!t]
\centering
\small
\setlength{\tabcolsep}{4pt}
\setlength{\belowcaptionskip}{4pt}
\caption{{Results on Cityscapes for semantic segmentation and depth estimation.}
ST baseline (Wan2.1) denotes single-task learning with the Wan2.1-T2V-14B backbone and VAE frozen.
\textbf{Bold}/\underline{underlined}: best/second best; $\uparrow$/$\downarrow$: higher/lower is better.}
\label{tab:cityscapes_results}
\begin{tabular*}{0.82\textwidth}{@{\extracolsep{\fill}}lcccccc@{}}
\toprule
\multirow{2}{*}{{Model}}
& \multicolumn{2}{c}{{Segmentation}} & \multicolumn{2}{c}{{Depth}}
& \multirow{2}{*}{{Params}} & \multirow{2}{*}{{${\Delta_m}$}} \\
\cmidrule(lr){2-3}
\cmidrule(lr){4-5}
& {mIoU} $\uparrow$ & {Pix Acc} $\uparrow$ & {Abs Err} $\downarrow$ & {Rel Err} $\downarrow$ &(M)$\downarrow$ &$[\%]$  $\uparrow$ \\
\midrule
HPS                                       & 67.40 & 90.92 & 0.0142 & 45.4262 & 55.76  & $0.00$  \\
STL                                       & 68.13 & 91.28 & 0.0133 & 45.0390 & 79.27  & $2.17$  \\
Cross-Stitch~\citep{misra2016crossstitch} & 68.01 & 91.29 & 0.0135 & 44.4246 & 79.27  & $2.11$  \\
MTAN~\citep{liu2019mtan}                  & 68.97 & 91.59 & 0.0136 & 43.7508 & 72.04  & $2.74$  \\
NDDR-CNN~\citep{gao2019nddrcnn}           & 68.02 & 91.25 & 0.0137 & 44.8662 & 101.58 & $1.51$  \\
VTAGML~\citep{bj2023VTAGML}               & 73.70 & 93.23 & 0.0138 & 42.8304 & 288.35 & $5.10$  \\
DenseMTL~\citep{lopes2023densemtl}        & 69.75 & 91.45 & 0.0152 & 52.1401 & 333.40 & $-4.44$ \\
SwinMTL~\citep{taghavi2024swinmtl}        & 73.33 & 92.87 & 0.0132 & 38.0720 & 333.31 & $8.54$  \\
LoRA-HPS                                  & 86.23 & 96.30 & 0.0123 & 34.2000 & \underline{37.35}  & $17.98$ \\
MultiLoRA~\citep{wang2023multilora}       & 87.23 & 96.67 & 0.0127 & \underline{31.0091} & 46.81  & $19.51$ \\
MTSAM~\citep{wang2025mtsam}               & \underline{87.45} & \underline{96.80} & \underline{0.0113} & 33.0086 & 37.44  & $\mathbf{20.99}$ \\
\hdashline
ST baseline (Wan2.1)                      & 79.32 & 94.44 & 0.0147 & 39.5952  &\textbf{6.60}   &0.00 \\
\method                                   & \textbf{88.91} & \textbf{97.17} & \textbf{0.0094} & \textbf{28.4684} & 108.16 & $\underline{19.78}$ \\
\bottomrule
\end{tabular*}
\end{table*}

\noindent
\textbf{Cityscapes performance.}
Table~\ref{tab:cityscapes_results} shows that \method{} achieves the best performance across all four metrics for semantic segmentation and depth estimation.
For semantic segmentation, \method{} reaches 88.91 mIoU and 97.17\% pixel accuracy, exceeding MTSAM~\citep{wang2025mtsam} by 1.46 and 0.37 points, respectively.
For depth estimation, it reduces absolute error by 16.81\% relative to MTSAM and relative error by 8.19\% relative to MultiLoRA~\citep{wang2023multilora}, the strongest competing baseline for this metric.
The comparison with the Wan2.1 ST baseline further shows consistent improvements across both tasks: segmentation mIoU increases by 9.59 points, while absolute and relative depth errors decrease by 36.05\% and 28.10\%, respectively.
Together, these improvements yield $\Delta_m=19.78\%$ relative to the Wan2.1 ST baseline.
Depth estimation contributes the larger share of this aggregate gain, accompanied by improvements in both segmentation metrics.
These results demonstrate the strength of \method{} in improving both semantic discrimination and geometric accuracy for urban scene understanding.

\begin{table}[!t]
\centering
\small
\setlength{\tabcolsep}{2.5pt}
\setlength{\belowcaptionskip}{3pt}
\caption{LoRA sharing on NYUD-v2 (Wan2.1-T2V-14B): one shared rank-16 adapter vs. one per task.
\textbf{Bold}/\underline{underlined}: best/second best; $\uparrow$/$\downarrow$: higher/lower is better.}
\label{tab:abl_shareLora}
\begin{tabular}{lccccc}
\toprule
\multirow{2}{*}{Setting}
& SemSeg & Depth & Normal & Bound  \\
& (mIoU)$\uparrow$ & (RMSE)$\downarrow$ & (mErr)$\downarrow$ & (odsF)$\uparrow$ \\
\midrule
Shared LoRA      &\underline{58.16}   &\textbf{0.3833}     &\underline{16.21}      &\underline{81.50}     \\
Task-routed LoRA &\textbf{60.91}   &\underline{0.3841}     &\textbf{15.95}      &\textbf{82.30}    \\
\bottomrule
\end{tabular}
\end{table}

\subsection{Ablation studies}

\noindent
\textbf{Ablation on Shared LoRA and Task-routed LoRA.}
Table~\ref{tab:abl_shareLora} compares shared and task-routed LoRA at rank $r=16$.
Task-routed LoRA improves semantic segmentation by 2.75 mIoU points and boundary detection by 0.80 odsF points, while reducing surface-normal error from 16.21 to 15.95.
Depth RMSE changes marginally, increasing from 0.3833 to 0.3841.
The comparison fixes the per-adapter rank, with task routing using four times the total LoRA parameters.
Under this setting, task-specific adaptation improves semantic parsing, normal estimation, and boundary localization with a small trade-off in depth accuracy.
These findings support task-specific specialization as an effective way to accommodate heterogeneous prediction requirements within a shared generative backbone.

\noindent
\textbf{Effect of LoRA rank.}
Table~\ref{tab:abl_loranum} shows that $r=32$ achieves the best segmentation and normal accuracy, with depth RMSE close to the best result at $r=8$. Increasing the rank to 64 improves boundaries but degrades the other three metrics while doubling adapter parameters.These results indicate that greater adaptation capacity does not uniformly improve performance across tasks. We therefore adopt $r=32$ for its strong semantic and geometric performance with fewer adapter parameters than $r=64$.

\begin{table}[!t]
\centering
\small
\caption{Task-routed LoRA rank on NYUD-v2 (default: $r=32$).
\textbf{Bold}/\underline{underlined}: best/second best; $\uparrow$/$\downarrow$: higher/lower is better.}
\label{tab:abl_loranum}
\begin{tabular}{lccccc}
\toprule
\multirow{2}{*}{Setting}
& SemSeg & Depth & Normal & Bound  \\
& (mIoU)$\uparrow$
& (RMSE)$\downarrow$
& (mErr)$\downarrow$
& (odsF)$\uparrow$  \\
\midrule
LoRA rank 8      &58.18  &\textbf{0.3800}   &16.19  &81.90 \\
LoRA rank 16     &\underline{60.91}  &0.3841   &15.95  &82.30  \\
LoRA rank 32     &\textbf{63.25}  &\underline{0.3815}   &\textbf{15.23}  &\underline{82.40} \\
LoRA rank 64     &59.23  &0.3876   &\underline{15.89}  &\textbf{83.30}  \\
\bottomrule
\end{tabular}
\end{table}

\begin{figure*}[!t]
    \centering
    \begin{minipage}[t]{0.585\textwidth}
        \centering
        \includegraphics[width=\linewidth]{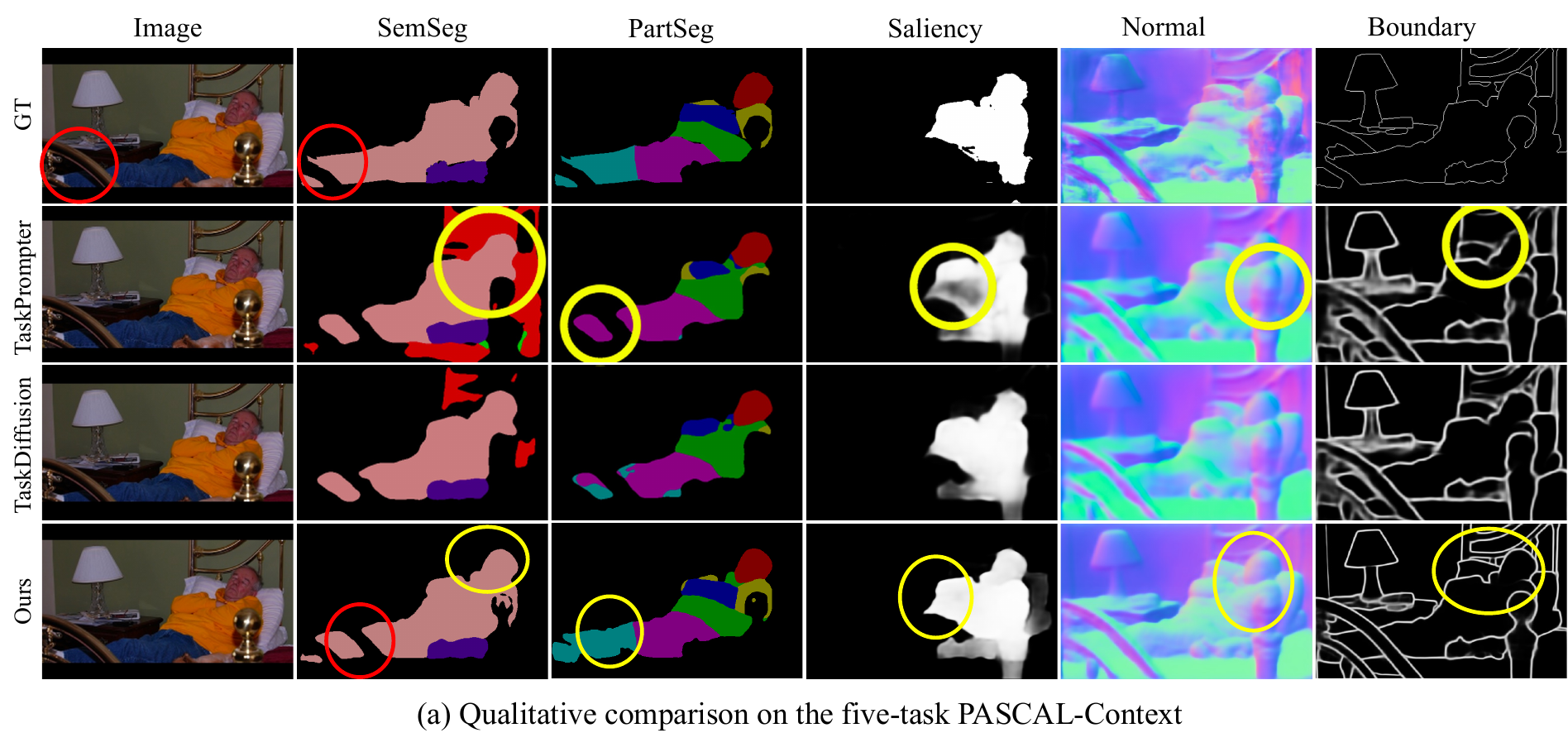}
    \end{minipage}
    \hfill
    \begin{minipage}[t]{0.41\textwidth}
        \centering
        \includegraphics[width=\linewidth]{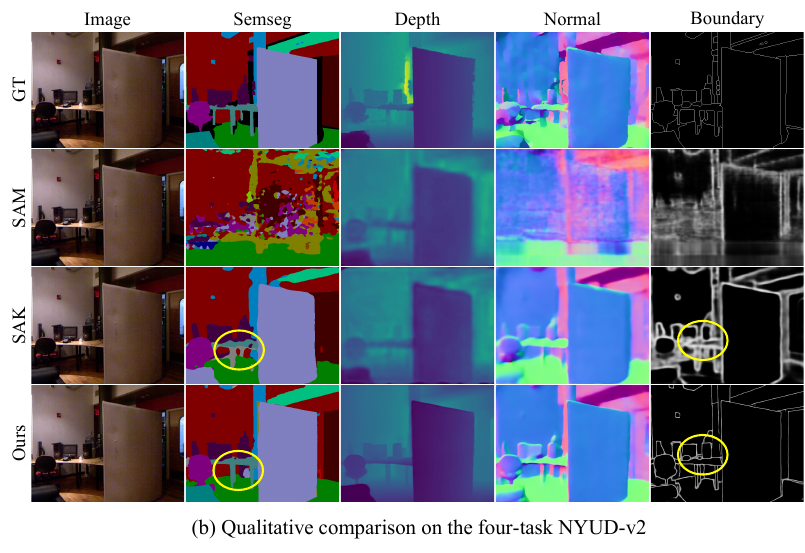}
    \end{minipage}

    \caption{
    (a) Qualitative comparison on the five-task PASCAL-Context benchmark with TaskPrompter~\citep{ye2023taskprompter} and TaskDiffusion~\citep{yang2025taskdiffusion}. Columns show the input image, semantic segmentation, human-part segmentation, saliency, surface normals, and boundary detection. (b) Qualitative comparison on the four-task NYUD-v2 benchmark with SAM~\citep{kirillov2023sam} and SAK~\citep{lu2025sak}.
    Columns show the input image, semantic segmentation, depth, surface normals, and boundaries.
    Rows show the ground truth and predictions
    from comparison methods and \method{}. 
    Red circles highlight regions where \method{}
    better captures image-visible occlusion structure.
    Yellow circles highlight improvements in
    semantic separation, saliency, boundary
    labeling, and structural detail.
    Best viewed in color and under magnification.
    }
    \label{fig:Qualitative}
\end{figure*}

\begin{table}[!t]
\centering
\small
\setlength{\tabcolsep}{1.5pt}
\renewcommand{\arraystretch}{1.05}
\setlength{\belowcaptionskip}{4pt}
\caption{Task-prompt ablation on NYUD-v2 with task-routed rank-16 LoRA. See Table~\ref{tab:prompt_definitions} for exact prompts.
  \textbf{Bold}/\underline{underlined}: best/second best; $\uparrow$/$\downarrow$: higher/lower is better.}
\label{tab:abl_prompts}

\begin{tabular}{lccccc}
\toprule
\multirow{2}{*}{Setting}
& SemSeg & Depth & Normal & Bound  \\
& (mIoU)$\uparrow$
& (RMSE)$\downarrow$
& (mErr)$\downarrow$
& (odsF)$\uparrow$\\
\midrule
 Task IDs                    &56.35    &0.3915   &16.32   &80.60 \\
 Task names                  &\underline{57.74}    &\underline{0.3829}   &\underline{15.14}   &\underline{82.20} \\
 Shared instruction          &56.95    &\textbf{0.3807}   &\textbf{15.00}   &\textbf{82.30}\\
 Full instructions~(\method) &\textbf{60.91}    &0.3841   &15.95   &\textbf{82.30}  \\
\bottomrule
\end{tabular}
\end{table}

\noindent
\textbf{Effect of task prompts.}
Table~\ref{tab:abl_prompts} compares four fixed text-conditioning schemes, retaining the correct adapter and prediction head for each task.
Task names outperform arbitrary task IDs across all four metrics.
Full instructions improve semantic segmentation to 60.91 mIoU, exceeding task names and the shared instruction by 3.17 and 3.96 points, respectively, and match the best boundary result of 82.30 odsF.
The shared instruction yields the lowest depth RMSE and normal error.
These results favor full instructions for semantic discrimination while showing textual conditioning affects the tasks differently within the shared generator.

\subsection{Visualization analysis}
Figure~\ref{fig:Qualitative}(a) compares \method{} with TaskPrompter~\citep{ye2023taskprompter} and TaskDiffusion~\citep{yang2025taskdiffusion} on the five-task PASCAL-Context benchmark.
The scene contains articulated body parts, foreground occlusions, and thin structures, presenting challenges for both semantic discrimination and spatial precision.
For semantic segmentation prediction, \method{} produces cleaner separation around the person's head.
For human-part segmentation prediction, it recovers more coherent lower-leg regions and clearer separation between upper- and lower-leg labels than the compared methods.
It also yields a more uniform saliency response within the torso region than TaskPrompter~\citep{ye2023taskprompter}, preserves local surface-orientation variations around the bedpost, and delineates thin bed-frame boundaries.
The red-circled region shows a localized discrepancy between the semantic prediction and the reference annotation around the partially occluded legs.
Here, the predicted regions follow the visible separation between the legs and the foreground bed rail.

Figure~\ref{fig:Qualitative}(b) compares \method{} with SAM~\citep{kirillov2023sam} and SAK~\citep{lu2025sak} on the four-task NYUD-v2 benchmark. \method{} preserves more coherent semantic regions and finer structural details. SAM produces fragmented segmentation and noisy surface normals, while SAK captures the overall layout but loses detail around the desk. In the highlighted regions, \method{} better separates the tabletop, thin supports, and background, with boundary contours more closely matching the ground truth. Its depth and surface-normal predictions also reveal clearer local geometry, suggesting improved preservation of scene structure across semantic and geometric tasks.
Appendix~\ref{app:visualizations} shows strong qualitative results across all three datasets, with some predictions appearing more faithful to visible image structures than the GT annotations.
These qualitative results highlight the promise of \method{} for adapting pretrained video generators into generalist multi-task dense predictors.

\section{Conclusions}
\label{sec:conclusion}

We present \method{}, a framework that adapts a pretrained video generator into a generalist backbone for multi-task dense prediction. Task-routed LoRA adapts the shared frozen backbone to individual tasks, while feature--latent fusion combines pre-projection transformer features with predicted latents. Task-specific heads map the fused representations to dense outputs and are trained directly with the corresponding task annotations. Each requested task requires a single deterministic DiT forward pass. Across PASCAL-Context, NYUD-v2, and Cityscapes, \method{} achieves leading results on diverse tasks and consistently improves over the corresponding Wan2.1 single-task baselines, with one model per benchmark. 
We believe our approach advances the use of pretrained video generators for multi-task dense prediction, combining broad task coverage with high prediction accuracy and opening new avenues for generalist visual understanding.


{    \small
    \bibliographystyle{ieeenat_fullname}
    \bibliography{main}

\begin{thebibliography}{54}
\providecommand{\natexlab}[1]{#1}
\providecommand{\url}[1]{\texttt{#1}}
\expandafter\ifx\csname urlstyle\endcsname\relax
  \providecommand{\doi}[1]{doi: #1}\else
  \providecommand{\doi}{doi: \begingroup \urlstyle{rm}\Url}\fi

\bibitem[Agiza et~al.(2024)Agiza, Neseem, and Reda]{agiza2024mtlora}
Ahmed Agiza, Marina Neseem, and Sherief Reda.
\newblock {MTLoRA}: Low-rank adaptation approach for efficient multi-task learning.
\newblock In \emph{CVPR}, pages 16196--16205, 2024.

\bibitem[Baranchuk et~al.(2022)Baranchuk, Rubachev, Voynov, Khrulkov, and Babenko]{baranchuk2022labels}
Dmitry Baranchuk, Ivan Rubachev, Andrey Voynov, Valentin Khrulkov, and Artem Babenko.
\newblock Label-efficient semantic segmentation with diffusion models.
\newblock In \emph{ICLR}, 2022.

\bibitem[Berman et~al.(2018)Berman, Triki, and Blaschko]{berman2018lovasz}
Maxim Berman, Amal~Rannen Triki, and Matthew~B. Blaschko.
\newblock The lov\'asz-softmax loss: A tractable surrogate for the optimization of the intersection-over-union measure in neural networks.
\newblock In \emph{CVPR}, 2018.

\bibitem[Bhattacharjee et~al.(2023)Bhattacharjee, S{\"u}sstrunk, and Salzmann]{bj2023VTAGML}
Deblina Bhattacharjee, Sabine S{\"u}sstrunk, and Mathieu Salzmann.
\newblock Vision transformer adapters for generalizable multitask learning.
\newblock In \emph{ICCV}, pages 18969--18980, 2023.

\bibitem[Br{\"u}ggemann et~al.(2021)Br{\"u}ggemann, Kanakis, Obukhov, Georgoulis, and Van~Gool]{bruggemann2021atrc}
David Br{\"u}ggemann, Menelaos Kanakis, Anton Obukhov, Stamatios Georgoulis, and Luc Van~Gool.
\newblock Exploring relational context for multi-task dense prediction.
\newblock In \emph{ICCV}, pages 15869--15878, 2021.

\bibitem[Cao et~al.(2026)Cao, Lopes, and de~Charette]{stablemtl2026}
Anh-Quan Cao, Ivan Lopes, and Raoul de Charette.
\newblock {StableMTL}: Repurposing latent diffusion models for multi-task learning from partially annotated synthetic datasets.
\newblock In \emph{CVPR}, 2026.

\bibitem[Chen et~al.(2022{\natexlab{a}})Chen, Saxena, Li, Fleet, and Hinton]{chen2022pix2seq}
Ting Chen, Saurabh Saxena, Lala Li, David~J. Fleet, and Geoffrey Hinton.
\newblock {Pix2Seq}: A language modeling framework for object detection.
\newblock In \emph{ICLR}, 2022{\natexlab{a}}.

\bibitem[Chen et~al.(2022{\natexlab{b}})Chen, Saxena, Li, Lin, Fleet, and Hinton]{chen2022pix2seqv2}
Ting Chen, Saurabh Saxena, Lala Li, Tsung-Yi Lin, David~J. Fleet, and Geoffrey~E. Hinton.
\newblock A unified sequence interface for vision tasks.
\newblock In \emph{NeurIPS}, pages 31333--31346, 2022{\natexlab{b}}.

\bibitem[Cordts et~al.(2016)Cordts, Omran, Ramos, Rehfeld, Enzweiler, Benenson, Franke, Roth, and Schiele]{cityscapes}
Marius Cordts, Mohamed Omran, Sebastian Ramos, Timo Rehfeld, Markus Enzweiler, Rodrigo Benenson, Uwe Franke, Stefan Roth, and Bernt Schiele.
\newblock The cityscapes dataset for semantic urban scene understanding.
\newblock In \emph{CVPR}, pages 3213--3223, 2016.

\bibitem[Fu et~al.(2024)Fu, Yin, Hu, Wang, Ma, Tan, Shen, Lin, and Long]{fu2024geowizard}
Xiao Fu, Wei Yin, Mu Hu, Kaixuan Wang, Yuexin Ma, Ping Tan, Shaojie Shen, Dahua Lin, and Xiaoxiao Long.
\newblock {GeoWizard}: Unleashing the diffusion priors for {3D} geometry estimation from a single image.
\newblock In \emph{ECCV}, pages 241--258, 2024.

\bibitem[Gabeur et~al.(2026)Gabeur, Long, Peng, Voigtlaender, Sun, Bao, Truong, Wang, Zhou, Barron, Genova, Kannen, Ben, Li, Guo, Yogin, Gu, Chen, Wang, Xie, Zhou, He, Funkhouser, Alayrac, and Soricut]{visionbanana2026}
Valentin Gabeur, Shangbang Long, Songyou Peng, Paul Voigtlaender, Shuyang Sun, Yanan Bao, Karen Truong, Zhicheng Wang, Wenlei Zhou, Jonathan~T Barron, Kyle Genova, Nithish Kannen, Sherry Ben, Yandong Li, Mandy Guo, Suhas Yogin, Yiming Gu, Huizhong Chen, Oliver Wang, Saining Xie, Howard Zhou, Kaiming He, Thomas Funkhouser, Jean-Baptiste Alayrac, and Radu Soricut.
\newblock Image generators are generalist vision learners.
\newblock \emph{arXiv preprint arXiv:2604.20329}, 2026.

\bibitem[Gao et~al.(2019)Gao, Ma, Zhao, Liu, and Yuille]{gao2019nddrcnn}
Yuan Gao, Jiayi Ma, Mingbo Zhao, Wei Liu, and Alan~L. Yuille.
\newblock {NDDR-CNN}: Layerwise feature fusing in multi-task {CNNs} by neural discriminative dimensionality reduction.
\newblock In \emph{CVPR}, pages 3205--3214, 2019.

\bibitem[He et~al.(2025)He, Li, Yin, Liang, Li, Zhou, Zhang, Liu, and Chen]{he2025lotus}
Jing He, Haodong Li, Wei Yin, Yixun Liang, Leheng Li, Kaiqiang Zhou, Hongbo Zhang, Bingbing Liu, and Yingcong Chen.
\newblock {Lotus}: Diffusion-based visual foundation model for high-quality dense prediction.
\newblock In \emph{ICLR}, 2025.

\bibitem[Hu et~al.(2022)Hu, Shen, Wallis, Allen-Zhu, Li, Wang, Wang, and Chen]{hu2022lora}
Edward~J. Hu, Yelong Shen, Phillip Wallis, Zeyuan Allen-Zhu, Yuanzhi Li, Shean Wang, Lu Wang, and Weizhu Chen.
\newblock {LoRA}: Low-rank adaptation of large language models.
\newblock In \emph{ICLR}, 2022.

\bibitem[Ji et~al.(2023)Ji, Chen, Xie, Hong, Liu, Liu, Lu, Li, and Luo]{ji2023ddp}
Yuanfeng Ji, Zhe Chen, Enze Xie, Lanqing Hong, Xihui Liu, Zhaoqiang Liu, Tong Lu, Zhenguo Li, and Ping Luo.
\newblock {DDP}: Diffusion model for dense visual prediction.
\newblock In \emph{ICCV}, pages 21741--21752, 2023.

\bibitem[Ke et~al.(2024)Ke, Obukhov, Huang, Metzger, Daudt, and Schindler]{ke2024marigold}
Bingxin Ke, Anton Obukhov, Shengyu Huang, Nando Metzger, Rodrigo~Caye Daudt, and Konrad Schindler.
\newblock Repurposing diffusion-based image generators for monocular depth estimation.
\newblock In \emph{CVPR}, pages 9492--9502, 2024.

\bibitem[Kirillov et~al.(2023)Kirillov, Mintun, Ravi, Mao, Rolland, Gustafson, Xiao, Whitehead, Berg, Lo, Doll{\'a}r, and Girshick]{kirillov2023sam}
Alexander Kirillov, Eric Mintun, Nikhila Ravi, Hanzi Mao, Chloe Rolland, Laura Gustafson, Tete Xiao, Spencer Whitehead, Alexander~C. Berg, Wan-Yen Lo, Piotr Doll{\'a}r, and Ross Girshick.
\newblock Segment anything.
\newblock In \emph{ICCV}, pages 4015--4026, 2023.

\bibitem[Kolesnikov et~al.(2022)Kolesnikov, Pinto, Beyer, Zhai, Harmsen, and Houlsby]{kolesnikov2022uvim}
Alexander Kolesnikov, Andr{\'e}~Susano Pinto, Lucas Beyer, Xiaohua Zhai, Jeremiah Harmsen, and Neil Houlsby.
\newblock {UViM}: A unified modeling approach for vision with learned guiding codes.
\newblock In \emph{NeurIPS}, pages 26295--26308, 2022.

\bibitem[Li et~al.(2023)Li, Zhu, Jiang, Zhu, Li, Yuan, Wang, Qiao, Wang, Wang, and Dai]{li2023uniperceiverv2}
Hao Li, Jinguo Zhu, Xiaohu Jiang, Xizhou Zhu, Hongsheng Li, Chun Yuan, Xiaohua Wang, Yu Qiao, Xiaogang Wang, Wenhai Wang, and Jifeng Dai.
\newblock {Uni-Perceiver v2}: A generalist model for large-scale vision and vision-language tasks.
\newblock In \emph{CVPR}, pages 2691--2700, 2023.

\bibitem[Liu et~al.(2019)Liu, Johns, and Davison]{liu2019mtan}
Shikun Liu, Edward Johns, and Andrew~J. Davison.
\newblock End-to-end multi-task learning with attention.
\newblock In \emph{CVPR}, pages 1871--1880, 2019.

\bibitem[Lopes et~al.(2023)Lopes, Vu, and De~Charette]{lopes2023densemtl}
Ivan Lopes, Tuan-Hung Vu, and Raoul De~Charette.
\newblock Cross-task attention mechanism for dense multi-task learning.
\newblock In \emph{WACV}, pages 2328--2337. IEEE, 2023.

\bibitem[Lu et~al.(2023)Lu, Clark, Zellers, Mottaghi, and Kembhavi]{lu2022unifiedio}
Jiasen Lu, Christopher Clark, Rowan Zellers, Roozbeh Mottaghi, and Aniruddha Kembhavi.
\newblock {Unified-IO}: A unified model for vision, language, and multi-modal tasks.
\newblock In \emph{ICLR}, 2023.

\bibitem[Lu et~al.(2024{\natexlab{a}})Lu, Clark, Lee, Zhang, Khosla, Marten, Hoiem, and Kembhavi]{lu2024unifiedio2}
Jiasen Lu, Christopher Clark, Sangho Lee, Zichen Zhang, Savya Khosla, Ryan Marten, Derek Hoiem, and Aniruddha Kembhavi.
\newblock {Unified-IO 2}: Scaling autoregressive multimodal models with vision language audio and action.
\newblock In \emph{CVPR}, pages 26439--26455, 2024{\natexlab{a}}.

\bibitem[Lu et~al.(2024{\natexlab{b}})Lu, Sirejiding, Ding, Wang, and Lu]{lu2023pgt}
Yuxiang Lu, Shalayiding Sirejiding, Yue Ding, Chunlin Wang, and Hongtao Lu.
\newblock {Prompt Guided Transformer} for multi-task dense prediction.
\newblock \emph{IEEE TMM}, 26:\penalty0 6375--6385, 2024{\natexlab{b}}.

\bibitem[Lu et~al.(2025)Lu, Cao, and Wang]{lu2025sak}
Yuxiang Lu, Shengcao Cao, and Yu-Xiong Wang.
\newblock Swiss army knife: Synergizing biases in knowledge from vision foundation models for multi-task learning.
\newblock In \emph{ICLR}, 2025.

\bibitem[Misra et~al.(2016)Misra, Shrivastava, Gupta, and Hebert]{misra2016crossstitch}
Ishan Misra, Abhinav Shrivastava, Abhinav Gupta, and Martial Hebert.
\newblock Cross-stitch networks for multi-task learning.
\newblock In \emph{CVPR}, pages 3994--4003, 2016.

\bibitem[Mottaghi et~al.(2014)Mottaghi, Chen, Liu, Cho, Lee, Fidler, Urtasun, and Yuille]{pascal2014}
Roozbeh Mottaghi, Xianjie Chen, Xiaobai Liu, Nam-Gyu Cho, Seong-Whan Lee, Sanja Fidler, Raquel Urtasun, and Alan Yuille.
\newblock The role of context for object detection and semantic segmentation in the wild.
\newblock In \emph{CVPR}, pages 891--898, 2014.

\bibitem[Nishi et~al.(2024)Nishi, Kim, Li, and Pfister]{nishi2024jointtask}
Kento Nishi, Junsik Kim, Wanhua Li, and Hanspeter Pfister.
\newblock Joint-task regularization for partially labeled multi-task learning.
\newblock In \emph{CVPR}, pages 16152--16162, 2024.

\bibitem[Ravishankar et~al.(2025)Ravishankar, Patel, Rajasegaran, and Malik]{ravishankar2025scaling}
Rahul Ravishankar, Zeeshan Patel, Jathushan Rajasegaran, and Jitendra Malik.
\newblock Scaling properties of diffusion models for perceptual tasks.
\newblock In \emph{CVPR}, pages 12945--12954, 2025.

\bibitem[Silberman et~al.(2012)Silberman, Hoiem, Kohli, and Fergus]{silberman2012nyud}
Nathan Silberman, Derek Hoiem, Pushmeet Kohli, and Rob Fergus.
\newblock Indoor segmentation and support inference from {RGBD} images.
\newblock In \emph{ECCV}, 2012.

\bibitem[Taghavi et~al.(2024)Taghavi, Langari, and Pandey]{taghavi2024swinmtl}
Pardis Taghavi, Reza Langari, and Gaurav Pandey.
\newblock Swinmtl: A shared architecture for simultaneous depth estimation and semantic segmentation from monocular camera images.
\newblock In \emph{IROS}, pages 4957--4964. IEEE, 2024.

\bibitem[Tang et~al.(2023)Tang, Jia, Wang, Phoo, and Hariharan]{wang2023dift}
Luming Tang, Menglin Jia, Qianqian Wang, Cheng~Perng Phoo, and Bharath Hariharan.
\newblock Emergent correspondence from image diffusion.
\newblock In \emph{NeurIPS}, 2023.

\bibitem[{Team Wan}(2025)]{wan2025}
{Team Wan}.
\newblock Wan: Open and advanced large-scale video generative models.
\newblock \emph{arXiv preprint arXiv:2503.20314}, 2025.

\bibitem[Vandenhende et~al.(2020)Vandenhende, Georgoulis, and Van~Gool]{vandenhende2020mtinet}
Simon Vandenhende, Stamatios Georgoulis, and Luc Van~Gool.
\newblock {MTI-Net}: Multi-scale task interaction networks for multi-task learning.
\newblock In \emph{ECCV}, pages 527--543, 2020.

\bibitem[Wang et~al.(2026)Wang, Zhang, Kabra, Uijlings, Waslander, Zisserman, Carreira, He, Andriluka, Bazavan, Zanfir, and Sminchisescu]{genception2026}
Letian Wang, Chuhan Zhang, Rishabh Kabra, Jasper Uijlings, Steven Waslander, Andrew Zisserman, Joao Carreira, Kaiming He, Misha Andriluka, Eduard~Gabriel Bazavan, Andrei Zanfir, and Cristian Sminchisescu.
\newblock Video generation models are general-purpose vision learners.
\newblock In \emph{ECCV}, 2026.

\bibitem[Wang et~al.(2022)Wang, Yang, Men, Lin, Bai, Li, Ma, Zhou, Zhou, and Yang]{wang2022ofa}
Peng Wang, An Yang, Rui Men, Junyang Lin, Shuai Bai, Zhikang Li, Jianxin Ma, Chang Zhou, Jingren Zhou, and Hongxia Yang.
\newblock {OFA}: Unifying architectures, tasks, and modalities through a simple sequence-to-sequence learning framework.
\newblock In \emph{ICML}, pages 23318--23340. PMLR, 2022.

\bibitem[Wang et~al.(2024)Wang, Li, Zhao, Lian, Huang, Wang, Li, and Gao]{wang2024tsptransformer}
Shuo Wang, Jing Li, Zibo Zhao, Dongze Lian, Binbin Huang, Xiaomei Wang, Zhengxin Li, and Shenghua Gao.
\newblock Tsp-transformer: Task-specific prompts boosted transformer for holistic scene understanding.
\newblock In \emph{WACV}, pages 925--934, 2024.

\bibitem[Wang et~al.(2023{\natexlab{a}})Wang, Wang, Cao, Shen, and Huang]{wang2023painter}
Xinlong Wang, Wen Wang, Yue Cao, Chunhua Shen, and Tiejun Huang.
\newblock Images speak in images: A generalist painter for in-context visual learning.
\newblock In \emph{CVPR}, 2023{\natexlab{a}}.

\bibitem[Wang et~al.(2025)Wang, Zhuang, Ye, and Zhang]{wang2025mtsam}
Xuehao Wang, Zhan Zhuang, Feiyang Ye, and Yu Zhang.
\newblock {MTSAM}: Multi-task fine-tuning for segment anything model.
\newblock In \emph{ICLR}, 2025.

\bibitem[Wang et~al.(2023{\natexlab{b}})Wang, Lin, Zeng, and Zhang]{wang2023multilora}
Yiming Wang, Yu Lin, Xiaodong Zeng, and Guannan Zhang.
\newblock Multilora: Democratizing lora for better multi-task learning.
\newblock \emph{arXiv preprint arXiv:2311.11501}, 2023{\natexlab{b}}.

\bibitem[Xu et~al.(2018)Xu, Ouyang, Wang, and Sebe]{xu2018padnet}
Dan Xu, Wanli Ouyang, Xiaogang Wang, and Nicu Sebe.
\newblock {PAD-Net}: Multi-tasks guided prediction-and-distillation network for simultaneous depth estimation and scene parsing.
\newblock In \emph{CVPR}, 2018.

\bibitem[Xu et~al.(2025)Xu, Ge, Liu, Fan, Xie, Zhao, Chen, and Shen]{xu2024genpercept}
Guangkai Xu, Yongtao Ge, Mingyu Liu, Chengxiang Fan, Kangyang Xie, Zhiyue Zhao, Hao Chen, and Chunhua Shen.
\newblock What matters when repurposing diffusion models for general dense perception tasks?
\newblock In \emph{ICLR}, 2025.

\bibitem[Xu et~al.(2023{\natexlab{a}})Xu, Yang, and Zhang]{xu2023demt}
Yangyang Xu, Yibo Yang, and Lefei Zhang.
\newblock {DeMT}: Deformable mixer transformer for multi-task learning of dense prediction.
\newblock In \emph{AAAI}, pages 3072--3080, 2023{\natexlab{a}}.

\bibitem[Xu et~al.(2023{\natexlab{b}})Xu, Yang, and Zhang]{xu2023mtkd}
Yangyang Xu, Yibo Yang, and Lefei Zhang.
\newblock Multi-task learning with knowledge distillation for dense prediction.
\newblock In \emph{ICCV}, pages 21550--21559, 2023{\natexlab{b}}.

\bibitem[Xu et~al.(2024)Xu, Li, Yuan, Yang, and Zhang]{xu2022mqtransformer}
Yangyang Xu, Xiangtai Li, Haobo Yuan, Yibo Yang, and Lefei Zhang.
\newblock Multi-task learning with {Multi-Query Transformer} for dense prediction.
\newblock \emph{IEEE TCSVT}, 34\penalty0 (2):\penalty0 1228--1240, 2024.

\bibitem[Yang et~al.(2025{\natexlab{a}})Yang, Qi, Li, Li, Jampani, and Yang]{yang2025unifieddensevideo}
Lehan Yang, Lu Qi, Xiangtai Li, Sheng Li, Varun Jampani, and Ming-Hsuan Yang.
\newblock Unified dense prediction of video diffusion.
\newblock In \emph{CVPR}, pages 28963--28973, 2025{\natexlab{a}}.

\bibitem[Yang et~al.(2024)Yang, Jiang, Hou, Zhang, Chen, and Li]{yang2024mlore}
Yuqi Yang, Peng-Tao Jiang, Qibin Hou, Hao Zhang, Jinwei Chen, and Bo Li.
\newblock Multi-task dense prediction via mixture of low-rank experts.
\newblock In \emph{CVPR}, pages 27927--27937, 2024.

\bibitem[Yang et~al.(2025{\natexlab{b}})Yang, Jiang, Hou, Zhang, Chen, and Li]{yang2025taskdiffusion}
Yuqi Yang, Peng-Tao Jiang, Qibin Hou, Hao Zhang, Jinwei Chen, and Bo Li.
\newblock Multi-task dense predictions via unleashing the power of diffusion.
\newblock In \emph{ICLR}, 2025{\natexlab{b}}.

\bibitem[Ye and Xu(2022)]{ye2022invpt}
Hanrong Ye and Dan Xu.
\newblock {InvPT}: Inverted pyramid multi-task transformer for dense scene understanding.
\newblock In \emph{ECCV}, pages 514--530, 2022.

\bibitem[Ye and Xu(2023{\natexlab{a}})]{ye2023taskexpert}
Hanrong Ye and Dan Xu.
\newblock {TaskExpert}: Dynamically assembling multi-task representations with memorial mixture-of-experts.
\newblock In \emph{ICCV}, pages 21828--21837, 2023{\natexlab{a}}.

\bibitem[Ye and Xu(2023{\natexlab{b}})]{ye2023taskprompter}
Hanrong Ye and Dan Xu.
\newblock {TaskPrompter}: Spatial-channel multi-task prompting for dense scene understanding.
\newblock In \emph{ICLR}, 2023{\natexlab{b}}.

\bibitem[Ye and Xu(2024)]{ye2024diffusionmtl}
Hanrong Ye and Dan Xu.
\newblock {DiffusionMTL}: Learning multi-task denoising diffusion model from partially annotated data.
\newblock In \emph{CVPR}, pages 27960--27969, 2024.

\bibitem[Zhao et~al.(2023)Zhao, Rao, Liu, Liu, Zhou, and Lu]{zhang2023vpd}
Wenliang Zhao, Yongming Rao, Zuyan Liu, Benlin Liu, Jie Zhou, and Jiwen Lu.
\newblock Unleashing text-to-image diffusion models for visual perception.
\newblock In \emph{ICCV}, 2023.

\bibitem[Zou et~al.(2023)Zou, Dou, Yang, Gan, Li, Li, Dai, Behl, Wang, Yuan, Peng, Wang, Lee, and Gao]{zou2023xdecoder}
Xueyan Zou, Zi-Yi Dou, Jianwei Yang, Zhe Gan, Linjie Li, Chunyuan Li, Xiyang Dai, Harkirat Behl, Jianfeng Wang, Lu Yuan, Nanyun Peng, Lijuan Wang, Yong~Jae Lee, and Jianfeng Gao.
\newblock Generalized decoding for pixel, image, and language.
\newblock In \emph{CVPR}, pages 15116--15127, 2023.

\end{thebibliography}
}


\clearpage
\appendix
\setcounter{page}{1}
\maketitlesupplementary

\section{Framework Details}
\label{app:framework}
\label{app:wan_architecture}

\paragraph{Why use Wan2.1 for dense image prediction?}

  Wan2.1-T2V-14B is pretrained first on text--image pairs and then jointly on images and videos~\citep{wan2025}. Its VAE encodes a single frame as a spatial grid of latent features, and its DiT processes that grid under text conditioning. 
  This combination lets us adapt a shared generative backbone to predict different properties at each image location, including semantic class, depth, and surface orientation.
  
  Given an RGB image, \method{} treats it as a single frame ($T=1$) and encodes it once with the frozen VAE. For each requested task $k$, we reuse the clean image latent, activate the corresponding LoRA adapter, and run a single DiT forward pass at $t=0$ conditioned on  the task instruction.

\subsection{Backbone and representation extraction.}
We detail the Wan2.1-T2V-14B backbone and the extraction of the pre-projection features $h_k$ and predicted latent $\hat z_k$ in Eq.~\ref{eq:formulation}.
We use the frozen VAE encoder, UMT5-XXL text encoder, and 40-block DiT of Wan2.1-T2V-14B~\citep{wan2025}.
The DiT has hidden width 5120, 40 attention heads, and feed-forward width 13824; text embeddings have width 4096.
For single images ($T=1$), the VAE posterior mode and pretrained channel-wise normalization produce a 16-channel latent at $1/8$ image resolution.
A $(1,2,2)$ patch embedding maps it to the transformer grid at $1/16$ image resolution.
The clean latent is processed at $t=0$ without added noise~\citep{genception2026}.

\begin{table}[h]
    \centering
    \small
    \caption{Tensor dimensions in Wan2.1-T2V-14B. Both branches share one task-conditioned forward pass; the 2D head folds the singleton frame dimension into the batch dimension.}
    \label{tab:wan_tensor_shapes}
    \footnotesize
    \setlength{\tabcolsep}{3pt}
    \begin{tabular}{lll}
        \toprule
        Tensor & Input $448\times576$ & Input $512\times512$ \\
        \midrule
        Input latent $z_x$ & $16\times1\times56\times72$ & $16\times1\times64\times64$ \\
        Pre-projection tokens $\tilde H_k$ & $1008\times5120$ & $1024\times5120$ \\
        Feature grid $h_k$ & $5120\times1\times28\times36$ & $5120\times1\times32\times32$ \\
        Projected tokens $V_k$ & $1008\times64$ & $1024\times64$ \\
        Predicted latent $\hat z_k$ & $16\times1\times56\times72$ & $16\times1\times64\times64$ \\
        Fused head features $f_k$ & $256\times56\times72$ & $256\times64\times64$ \\
        \bottomrule
    \end{tabular}
\end{table}

A forward pre-hook on \texttt{transformer.proj\_out} captures the normalized, timestep-modulated tokens without detaching them.
Reshaping these tokens gives $h_k$; projecting and unpatchifying them gives the velocity, whose negative is $\hat z_k$ (Eq.~\ref{eq:one_step}).
Both representations come from the same task-conditioned forward pass.
Table~\ref{tab:wan_tensor_shapes} gives their dimensions.

\paragraph{Task adapters and prediction heads.}
Each task uses an independent rank-32 LoRA with $\alpha=16$ and zero dropout on the query, key, value, and output projections of self-attention and cross-attention in all 40 blocks.
The known task identity selects the adapter and its prediction head.
All original DiT, VAE, and text-encoder weights remain frozen.
The base head (Table~\ref{tab:app_decoder}) projects $h_k$ and $\hat z_k$ to 256 channels and aligns their spatial sizes before additive fusion.
Each upsampling block uses bilinear interpolation with \texttt{align\_corners=False}, depthwise $3\times3$ and pointwise convolutions, eight-group normalization, GELU, and residual refinement.
The output width $C_k$ is 40/21/7 for NYUD-v2/PASCAL-Context/Cityscapes semantic segmentation, seven for human parts, two for saliency, three for normals, and one for depth or boundaries.
\begin{table}[t]
\centering
\small
\caption{Base feature--latent decoder. Spatial sizes are relative to the input image. Each task has independent decoder parameters.}
\label{tab:app_decoder}
\footnotesize
\setlength{\tabcolsep}{2.5pt}
\begin{tabular}{@{}lll@{}}
\toprule
Operation & Channels & Spatial size \\
\midrule
Feature projection: $1\times1$, normalization, GELU & $5120\rightarrow256$ & $H/16\times W/16$ \\
Feature alignment: upsampling and refinement & $256\rightarrow256$ & $H/8\times W/8$ \\
Latent projection: $1\times1$, normalization & $16\rightarrow256$ & $H/8\times W/8$ \\
Addition, GELU, residual refinement & $256$ & $H/8\times W/8$ \\
Upsampling block 1 & $256\rightarrow128$ & $H/4\times W/4$ \\
Upsampling block 2 & $128\rightarrow64$ & $H/2\times W/2$ \\
Upsampling block 3 & $64\rightarrow32$ & $H\times W$ \\
Task output: $1\times1$ convolution & $32\rightarrow C_k$ & $H\times W$ \\
\bottomrule
\end{tabular}
\end{table}
The refined NYUD-v2 semantic head additionally uses a quarter-resolution RGB stem with channels $3\rightarrow64\rightarrow128$ and features from DiT blocks 9 and 24 (zero-based).
The RGB and intermediate-feature branches enter through learned scalar gates initialized to zero.
Other task heads and all controlled ablations retain the base decoder.

\paragraph{Task-specific losses.}
  \label{app:task_losses}
All losses use valid pixels, ignoring label 255 for categorical tasks.
  Semantic segmentation uses cross-entropy (CE) plus $0.5$ Lov\'asz--Softmax~\citep{berman2018lovasz}, human parts use $2\,\mathrm{CE}$, and saliency uses $5\times$ class-balanced CE.
  NYUD-v2 uses $\mathcal{L}_1+0.1\mathcal{L}_{\mathrm{SiLog}}$ for depth and $5\times$ cosine loss plus $2\times$ angular error for normals.
  Depth and Normals both use $10\mathcal{L}_1$. Boundary uses $50\times$ balanced binary CE.

\begin{table*}[t]
\centering
\small
\caption{Full text of the task-specific instructions during training and inference on NYUD-v2, PASCAL-Context, and Cityscapes.}
\label{tab:app_prompts}
\begin{tabular}{@{}llp{0.78015\linewidth}@{}}
\toprule
Dataset & Task & Instruction \\
\midrule
NYUD-v2 & SemSeg & Predict the 40-class semantic segmentation of this indoor scene. \\
NYUD-v2 & Depth & Estimate the dense monocular depth map of this indoor scene. \\
NYUD-v2 & Normal & Estimate the per-pixel surface normals of this indoor scene. \\
NYUD-v2 & Boundary & Detect the per-pixel semantic object boundaries of this indoor scene. \\
\midrule
PASCAL-Context & SemSeg & Predict the 21-class semantic segmentation of this PASCAL-Context scene at every pixel. \\
PASCAL-Context & Parts & Segment the person into six anatomical part classes plus background at every pixel. \\
PASCAL-Context & Saliency & Predict the binary salient foreground region of this scene at every pixel. \\
PASCAL-Context & Normal & Estimate the three-dimensional unit surface normal vector at every valid pixel. \\
PASCAL-Context & Boundary & Detect only thin one-pixel semantic object boundaries for this scene. \\
 \midrule
  Cityscapes & SemSeg & Predict the seven-class semantic segmentation of this urban street scene, including flat surfaces, construction, objects, nature, sky, humans,
  and vehicles. \\
  Cityscapes & Depth & Predict the dense per-pixel depth target of this urban street scene in the provided dataset representation. \\
\bottomrule
\end{tabular}
\end{table*}

 \begin{table*}[!t]
      \centering
      \small
      \caption{Exact wording of the task IDs, task names, and full instructions compared in the NYUD-v2 prompt ablation.}
      \label{tab:prompt_definitions}
      \begin{tabular}{lllll}
          \toprule
          Task & Task ID & Task name & Full instruction \\
          \midrule
          SemSeg & Task 0 & Semantic segmentation &
          Predict the 40-class semantic segmentation of this indoor scene. \\
          Depth & Task 1 & Depth estimation &
          Estimate the dense monocular depth map of this indoor scene. \\
          Normal & Task 2 & Surface normal estimation &
          Estimate the per-pixel surface normals of this indoor scene. \\
          Boundary & Task 3 & Semantic boundary detection &
          Detect the per-pixel semantic object boundaries of this indoor scene. \\
          \bottomrule
      \end{tabular}
  \end{table*}

\section{Experimental Details}
\label{app:training}
\label{app:experiments}

\subsection{Implementation Details}
\label{app:implementation}
\label{app:evaluation}
 
For initial joint training, we use Adam with a 500-update linear warm-up and polynomial decay (power $0.9$) on a 40k-update schedule.
  We report base learning rates as (LoRA adapter, decoder) pairs in units of $10^{-4}$, where the decoder includes feature fusion and prediction.
  The adapters are frozen for the first 1,000 updates on NYUD-v2 and PASCAL-Context and the first 500 on Cityscapes.
  We clip the adapter and decoder gradient norms at $1$ and $10$, respectively. Weight decay is $10^{-6}$ on PASCAL-Context and zero on NYUD-v2 and Cityscapes.

\paragraph{Task prompts.}
  \label{app:prompts}
  Table~\ref{tab:app_prompts} lists the full instruction that \method{} uses for each dataset and task.
  We encode each prompt once with the frozen UMT5-XXL encoder and reuse its padded $226\times4096$ embedding across images. The same task instruction is used for all images of a given benchmark during both training and evaluation.
  Table~\ref{tab:prompt_definitions} gives the three task-specific prompt formats in the NYUD-v2 ablation.
  For semantic segmentation, \texttt{Task 0} is an arbitrary identifier, \texttt{Semantic segmentation} names the task, and the full instruction specifies the 40-class label space and the indoor setting.
  In the shared-instruction control, all four tasks receive ``Infer semantic regions, metric depth, surface normals, and object boundaries for this indoor scene.''
  In every variant, the task identity selects the corresponding LoRA adapter and prediction head.

\section{ Additional Experiments}
\label{app:additional_results}

\subsection{Ablation on LoRA sharing}
At rank 16, using one LoRA adapter per task improves SemSeg mIoU by 2.75 points over a single shared adapter, with four times as many LoRA parameters (Table~\ref{tab:app_lora_sharing}).
At the same total budget of 209.72M LoRA parameters, the four task-specific adapters achieve 60.91 mIoU and 82.30 boundary ODS, compared with 59.46 and 82.10 for the shared rank-64 adapter.
Task routing also reduces mean surface-normal error from $16.14$ to $15.95$, while depth RMSE increases slightly from 0.3822 to 0.3841.
The routed model activates 52.43M LoRA parameters, one quarter of the 209.72M used by the shared rank-64 adapter.

\begin{table}[t]
\centering
\small
\setlength{\tabcolsep}{3pt}
\caption{LoRA sharing ablation on NYUD-v2 at 20k updates. Parameter counts include LoRA weights only; task-routed LoRA uses four adapters.
Best/second-best values are \textbf{bold}/\underline{underlined}; $\uparrow$/$\downarrow$: higher/lower is better.}
\label{tab:app_lora_sharing}
\resizebox{\columnwidth}{!}{%
\begin{tabular}{@{}lcrrrr@{}}
\toprule
Setting & \shortstack{LoRA\\Params (M)}
& \shortstack{SemSeg\\mIoU $\uparrow$}
& \shortstack{Depth\\RMSE $\downarrow$}
& \shortstack{Normal\\mErr $\downarrow$}
& \shortstack{Bound\\ODS $\uparrow$} \\
\midrule
Shared ($r=16$)           & \textbf{52.43}   & 58.16  & \underline{0.3833}  & 16.21  & 81.5   \\
Shared ($r=64$)           & \underline{209.72}  & \underline{59.46}  & \textbf{0.3822}  & \underline{16.14}  & \underline{82.1} \\
Task-routed ($4\times16$) & \underline{209.72}  & \textbf{60.91}  & 0.3841  & \textbf{15.95}  & \textbf{82.3} \\
\bottomrule
\end{tabular}%
}
\end{table}

\subsection{Ablation on backbone comparison}

At a fixed LoRA rank of 16, Wan2.1-T2V-14B improves on its 1.3B counterpart across all four NYUD-v2 tasks (Table~\ref{tab:app_backbone}). The gain is particularly pronounced for semantic segmentation, where mIoU rises from 49.38 to 60.91, an increase of 11.53 points. 
The two larger backbones exhibit a trade-off between semantic segmentation and surface-normal estimation. Wan2.1-T2V-14B leads Wan2.2-T2V-A14B by 2.35 mIoU points (60.91 vs.\ 58.56), while Wan2.2 achieves the lowest mean surface-normal error ($15.00$ vs.\ $15.95$). 
These results support Wan2.1-T2V-14B as the default backbone for \method{}, as it achieves the best reported performance on three of the four tasks.

\begin{table}[!t]
  \centering
  \small
  \setlength{\tabcolsep}{3pt}
  \caption{Comparison of Wan2.1 and Wan2.2 backbones across four dense prediction tasks on NYUD-v2.
  Each task uses its own rank-16 LoRA adapter.
  For metrics with at least two reported values, we highlight the \textbf{best} and \underline{second-best} results.}
  \label{tab:app_backbone}
  \begin{tabular}{@{}lcccc@{}}
  \toprule
  \multirow{2}{*}{Decoder input} & SemSeg & Depth & Normal & Bound \\
  & (mIoU)$\uparrow$ & (RMSE)$\downarrow$ & (mErr)$\downarrow$ & (odsF)$\uparrow$ \\
  \midrule
Wan2.1-T2V-1.3B          &49.38     & 0.4200             & 16.40             & 81.7\\
Wan2.1-T2V-14B  & \textbf{60.91}    & \textbf{0.3841}    & \underline{15.95}    & \textbf{82.3} \\
Wan2.2-T2V-A14B & \underline{58.56} & \underline{0.3852} & \textbf{15.00} &82.1 \\
  \bottomrule
  \end{tabular}
  \end{table}

\begin{algorithm}[t]
  \caption{GenMTL training}
  \label{alg:genmtl_train}
  \begin{algorithmic}[1]
  \Require Task set $\mathcal K$, training data, and task instructions $\{p_k\}$
  \State Freeze the VAE encoder $E$, text encoder $T_{\mathrm{text}}$, and DiT base weights $\theta_0$
  \State Cache $c_k \gets T_{\mathrm{text}}(p_k)$ for each $k\in\mathcal K$
  \State Initialize the LoRA, fusion, and head parameters
         $\Phi=\{(\Delta\theta_k,\eta_k,\psi_k)\}_{k\in\mathcal K}$
  \For{each training update}
      \ForAll{$k\in\mathcal K$}
          \State Load a task batch $(x_k,y_k,\Omega_k)$
          \State $\hat y_k \gets$ \Call{PredictTask}{$E(x_k),k$}
          \State $L_k \gets \ell_k(\hat y_k,y_k;\Omega_k)$
      \EndFor
      \State Update $\Phi$ using the mean loss
             $|\mathcal K|^{-1}\sum_{k\in\mathcal K}L_k$
  \EndFor
  \State \Return $\Phi$
  \end{algorithmic}
  \end{algorithm}

\subsection{Ablation on feature-latent fusion}

Table~\ref{tab:app_featureLatent} shows that the task heads benefit from direct access to the final pre-projection DiT features $h_k$ alongside the predicted latent $\hat z_k$.
With the same rank-16 LoRA configuration, fusion raises semantic segmentation mIoU from 58.67 to 60.91, a gain of 2.24 points. It also reduces depth RMSE from 0.3913 to 0.3841, a relative reduction of 1.84\%.
These gains suggest that $h_k$ complements the latent for both assigning semantic labels and estimating depth, two tasks that require different information at each pixel.

\section{Training and inference.}
  Algorithm~\ref{alg:genmtl_train} trains the task-specific LoRA adapters, fusion modules, and prediction heads by minimizing the mean of the masked task losses. The pretrained encoders and DiT base weights remain frozen.
  At inference, Algorithm~\ref{alg:genmtl_inference} encodes the image once and reuses its latent across all requested tasks.
  Each task uses its own adapter and cached instruction embedding for one DiT pass at $t=0$, then fuses the predicted latent with the final pre-projection features to produce a dense prediction.

\begin{table}[!t]
  \centering
  \small
  \setlength{\tabcolsep}{3pt}
  \caption{Ablation of feature--latent fusion on NYUD-v2.
  The fusion module combines the predicted latent $\hat z_k$ with the final pre-projection DiT features $h_k$.
  The latent-only variant sets $h_k$ to zero before fusion.
  Both variants use one rank-16 LoRA adapter per task.
  \textbf{Bold} indicates the better result for each metric.}   
  \label{tab:app_featureLatent}
  \begin{tabular}{@{}lcccc@{}}
  \toprule
  \multirow{2}{*}{Decoder input} & SemSeg & Depth & Normal & Bound \\
  & (mIoU)$\uparrow$ & (RMSE)$\downarrow$ & (mErr)$\downarrow$ & (odsF)$\uparrow$ \\
  \midrule
  Latent only           &        {58.67}         & 0.3913          & {15.97}                  & 82.0 \\
  Latent + DiT feature  & \textbf{60.91}         & \textbf{0.3841} & \textbf{15.95}           & \textbf{82.3} \\
  \bottomrule
  \end{tabular}
  \end{table}

  \begin{algorithm}[t!]
  \caption{GenMTL inference}
  \label{alg:genmtl_inference}
  \begin{algorithmic}[1]
  \Require Image $x$, requested tasks $\mathcal S\subseteq\mathcal K$,
           trained parameters $\Phi$, and cached embeddings $\{c_k\}$
  \Function{PredictTask}{$z,k$}
      \State $(\hat z_k,h_k) \gets F_{\theta_0,\Delta\theta_k}(z,c_k;t=0)$
             \Comment{$\hat z_k=-v_k$}
      \State $f_k \gets M_{\eta_k}(\hat z_k,h_k)$
      \State \Return $\rho_k\!\left(\operatorname{Resize}_k(D_{\psi_k}(f_k))\right)$
  \EndFunction
  \State $z_x \gets E(x)$
  \ForAll{$k\in\mathcal S$}
      \State $\hat y_k \gets$ \Call{PredictTask}{$z_x,k$}
  \EndFor
  \State \Return $\{\hat y_k\}_{k\in\mathcal S}$
  \end{algorithmic}
  \end{algorithm}

\noindent
 \textbf{Inference cost.}
  GenMTL encodes each input view once with the frozen VAE and uses a single Wan2.1-T2V-14B DiT forward pass per requested task.
  We measure end-to-end latency on a single NVIDIA A40 at batch size 1, averaging over 30 PASCAL-Context validation images after five warm-up images.
  At $512\times512$, single-view semantic segmentation takes 0.50\,s per image (1.98 images/s), including 0.43\,s in the DiT.
  Predicting all five tasks takes 2.30\,s per image (0.43 images/s), with 2.17\,s spent across the five DiT passes.
  Both settings peak at 28.13\,GiB of allocated GPU memory.
  The VAE weights occupy 0.24\,GiB, and the five cached task embeddings occupy 8.83\,MiB.

\begin{figure*}[!t!]
\centering
  \includegraphics[width=0.9299\textwidth]{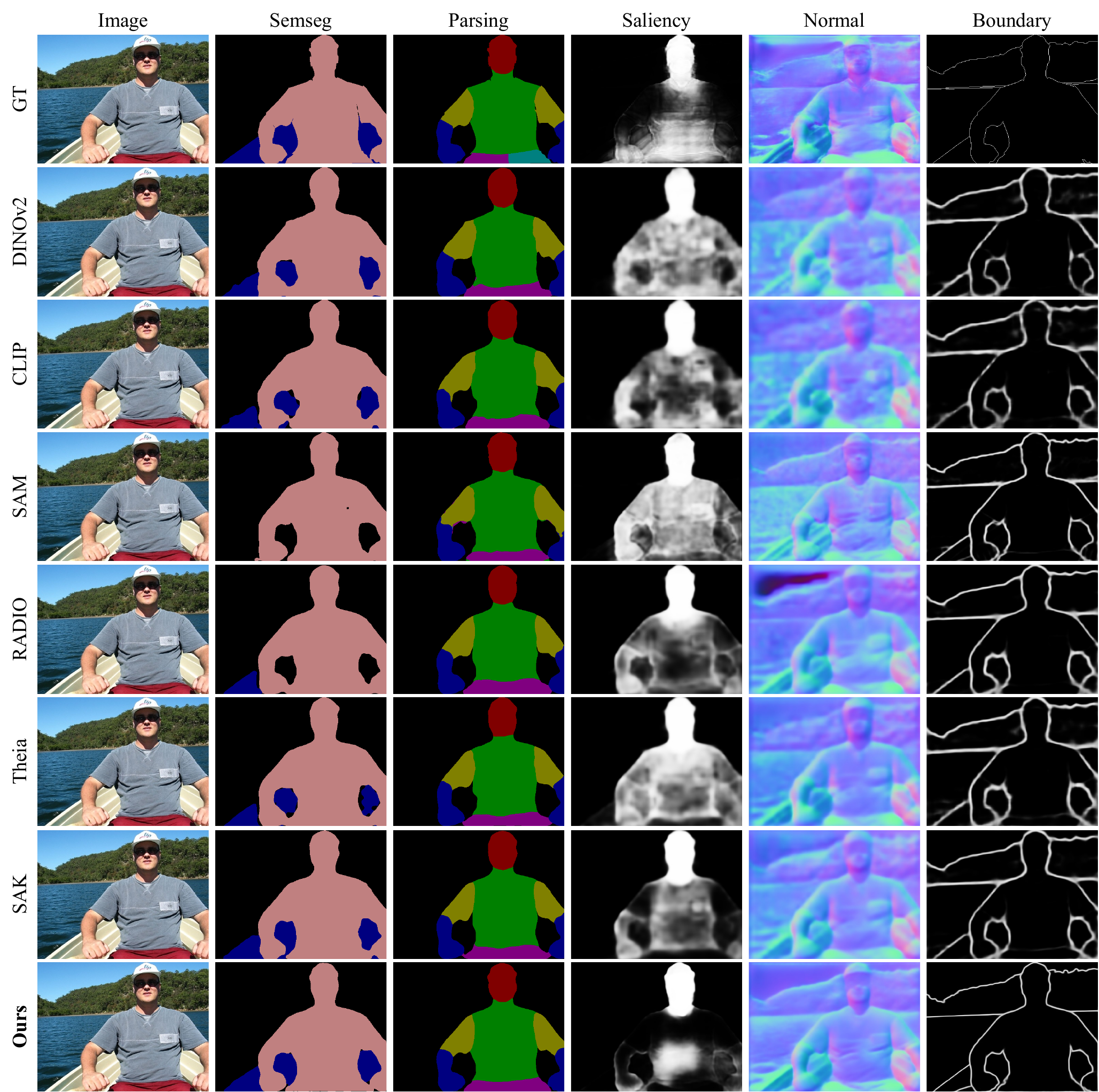}
      \caption{Five-task qualitative comparison on PASCAL-Context.
  From left to right, the columns show the RGB image, semantic segmentation, human-part segmentation, saliency, surface normals, and boundaries.
  The top row shows the reference annotations, followed by predictions from DINOv2, CLIP, SAM, RADIO, Theia, SAK, and \method{}.
  In the two gaps between the person's arms and torso, \method{} and SAK identify the boat, while SAM and RADIO predict background.}
\label{fig:app_psacal_vis}
\end{figure*}

\section{More Visualizations}
  \label{app:visualizations}
 
\subsection{Additional Qualitative Comparisons}
The methods in Figure~\ref{fig:app_psacal_vis} agree on the seated person's silhouette. Their semantic predictions differ in the gaps between the bent arms and torso, where \method{} and SAK~\citep{lu2025sak} identify the exposed boat. SAM~\citep{kirillov2023sam} and RADIO label both gaps as background. Along the arms and shoreline, \method{} produces narrower boundary responses than DINOv2 and CLIP. Its contour along the forearm on the right of the image follows the person--boat boundary visible in both the image and the semantic mask. The boundary annotation marks only a short segment near the hand.

Within the silhouette, the part map from \method{} separates upper arms from forearms near the sleeve hems, and the predicted surface normals capture the curvature of the elbows and several folds across the shirt.
Below the shirt, the part map merges two regions that the reference labels separately.
Both saliency maps emphasize the face. On the torso, \method{} highlights a compact region near the center of the shirt, while the reference response extends across the lower torso to the waist.

\begin{figure}[!t]
\centering
  \includegraphics[width=0.48999\textwidth]{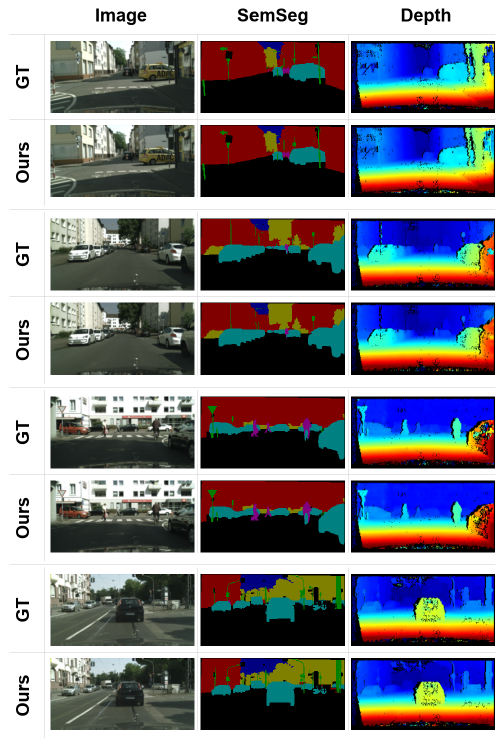}
    \caption{Qualitative results on Cityscapes. For each frame, GT is shown above Ours. Columns show the input image, semantic segmentation, and depth. The scenes cover street corners, parked vehicles, pedestrians and cyclists, and varying road depth.}
\label{fig:app_cityscapes_qualitative}
\end{figure}

\subsection{Additional visualizations}
\begin{figure*}[!t]
\centering
  \includegraphics[width=0.930\textwidth]{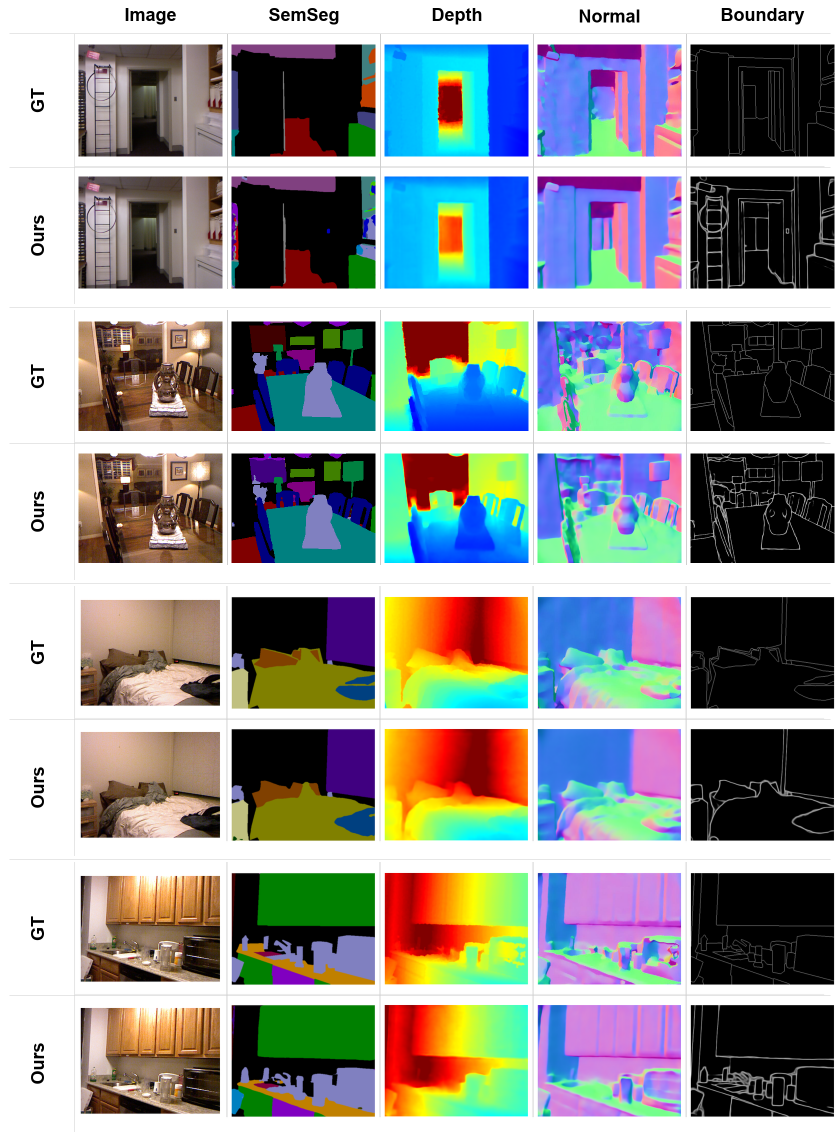}
    \caption{Qualitative results on NYUD-v2. For each sample, GT is shown above Ours. Columns show the input image, semantic segmentation, depth, surface normals, and boundaries. The scenes depict a corridor-like doorway, a long-table dining room, a bedroom, and a kitchen, respectively.}
\label{fig:app_nyud_qualitative}
\end{figure*}

\begin{figure*}[t!]
\centering
  \includegraphics[width=0.930\textwidth]{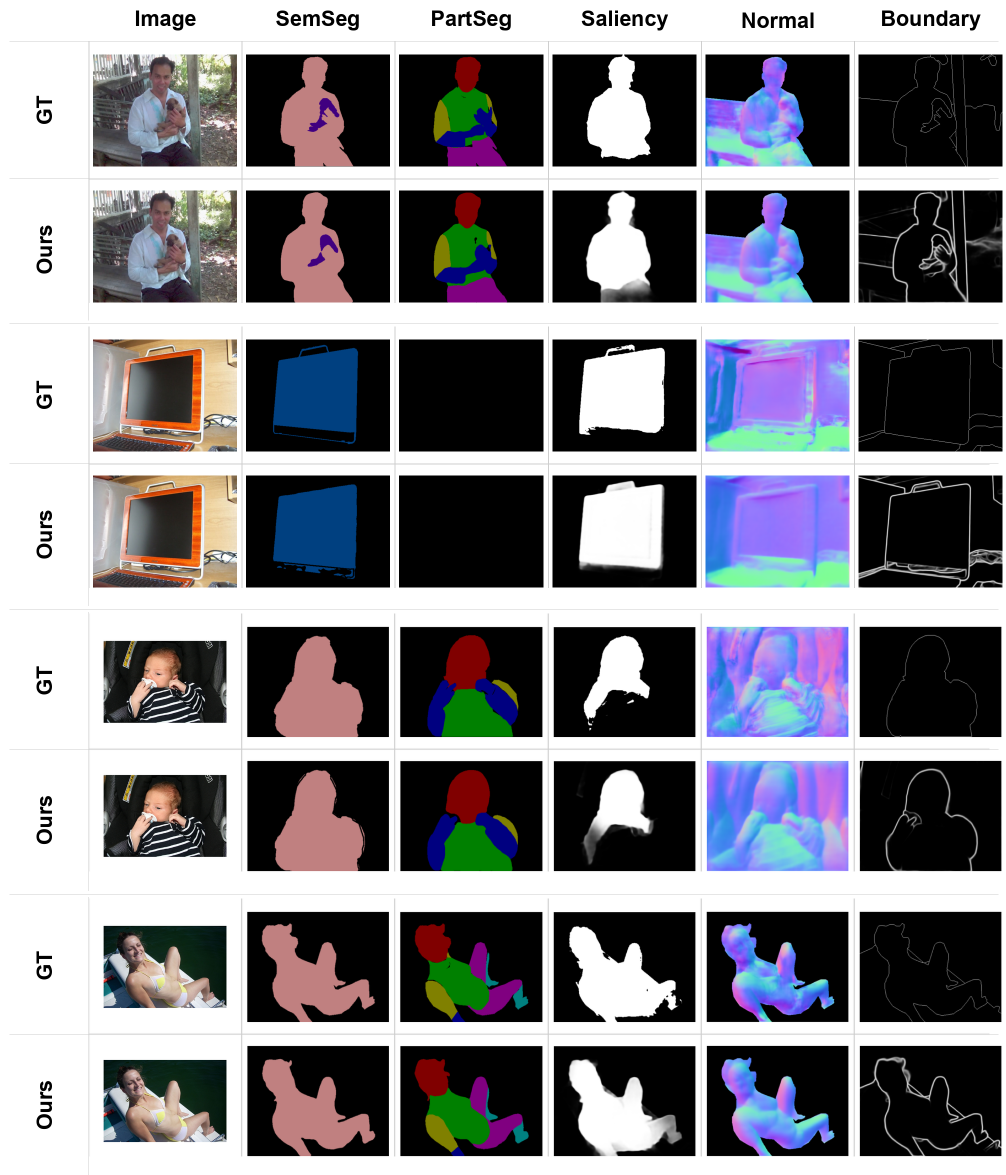}
\caption{Qualitative results on PASCAL-Context. For each sample, GT is shown above Ours. Columns show the input image, semantic segmentation, human-part segmentation, saliency, surface normals, and boundaries.}
\label{fig:app_pascal_qualitative}
\end{figure*}

Figures~\ref{fig:app_cityscapes_qualitative}, \ref{fig:app_nyud_qualitative}, and~\ref{fig:app_pascal_qualitative} show qualitative results on Cityscapes, NYUD-v2, and PASCAL-Context, respectively.
For each RGB image, the upper row shows the reference maps and the lower row shows the predictions from \method{}, with corresponding tasks aligned by column.
In the street and indoor scenes, we examine whether depth transitions follow the contours of occluding objects and how surface orientation varies within a semantic region on NYUD-v2.
For the PASCAL-Context examples, we compare each person's silhouette with the part labels inside it and examine which regions receive the strongest saliency response.
Across these three datasets, \method{} illustrates how a generative model can serve as a generalist multi-task dense predictor.

\paragraph{Cityscapes.}
  Figure~\ref{fig:app_cityscapes_qualitative} pairs semantic segmentation with depth in street scenes. Along the receding road, depth varies within a single semantic class. Where a parked vehicle occludes the street behind it, its silhouette marks both a class boundary
  and a depth discontinuity. Comparing the two predictions with their references helps distinguish a misplaced vehicle mask from a depth estimate that blends the vehicle into the background.
  Contour placement is particularly important for pedestrians and riders. Across all 500 validation images, roughly three quarters of the annotated \emph{human} pixels lie in a two-pixel band just inside the class boundary. With horizontal-flip test-time augmentation,
  \method{} correctly labels 86.15\% of the pixels in this band and 99.26\% of those in the mask interiors.

\paragraph{NYUD-v2.}
  At the doorway in Figure~\ref{fig:app_nyud_qualitative}, the frame occludes surfaces farther into the room. A depth prediction can preserve this ordering even if it blurs the transition across the frame. Comparing the transition with the annotated contour helps assess how precisely \method{} localizes the occluding edge.
  The long dining table illustrates depth variation within a single object. Depth increases from the near edge to the far end, while the planar tabletop should have nearly uniform surface normals.
  In the bedroom, the annotated bed silhouette lets us assess both the coverage of the semantic mask and the continuity of the predicted outline. In the kitchen, two adjoining faces of the same cabinet can have sharply different normals while sharing a semantic label.
 Comparing the doorway and the cabinet gives a direct check on the predicted depth map, which should show a sharp jump at the occluding frame and remain continuous where the two cabinet faces meet.

\paragraph{PASCAL-Context.}
For the seated person on the boat, as shown in Figure~\ref{fig:app_pascal_qualitative}, \method{} preserves the narrow gap between the raised foot and the opposite thigh in both the semantic mask and the boundary map. The part map distinguishes the upper and lower legs at the knees. Across the thighs, the predicted surface normals capture the overall curvature but show less local variation than the distilled reference. In the close-up of the child, the person mask includes the striped shirt, while both the predicted and reference saliency maps emphasize the head and raised hands. The boundary prediction traces the head and shoulders and adds a short contour where the fingers meet the face. This contour follows the hand--head division in the part annotation. The reference boundary map traces only the outer silhouette in this region.
 The hand--face contour illustrates how an anatomically accurate part boundary can reduce precision when evaluated against the object-level silhouette.

\end{document}